\documentclass[lettersize,journal]{IEEEtran}
\usepackage{amsmath,amsfonts}
\usepackage{algorithmic}
\usepackage{algorithm}
\usepackage{array}
\usepackage[caption=false,font=normalsize,labelfont=sf,textfont=sf]{subfig}
\usepackage{textcomp}
\usepackage{stfloats}
\usepackage{url}
\usepackage{verbatim}
\usepackage{graphicx}
\usepackage{multirow}
\usepackage{amssymb}
\usepackage{amsthm}
\usepackage{booktabs}
\usepackage[table]{xcolor}
\usepackage[numbers]{natbib}
\usepackage[pagebackref,breaklinks,colorlinks]{hyperref}
\usepackage{listings}
\usepackage[table]{xcolor}
\usepackage[numbers]{natbib}
\usepackage[pagebackref,breaklinks,colorlinks]{hyperref}
\usepackage{marvosym}
\usepackage{enumitem}
\lstdefinelanguage{json}{
    basicstyle=\ttfamily\small,
    numbers=left,
    numberstyle=\tiny\color{gray},
    stepnumber=1,
    numbersep=5pt,
    showstringspaces=false,
    breaklines=true,
    frame=single,
    backgroundcolor=\color{white},
    literate=
     *{0}{{{\color{blue}0}}}{1}
      {1}{{{\color{blue}1}}}{1}
      {2}{{{\color{blue}2}}}{1}
      {3}{{{\color{blue}3}}}{1}
      {4}{{{\color{blue}4}}}{1}
      {5}{{{\color{blue}5}}}{1}
      {6}{{{\color{blue}6}}}{1}
      {7}{{{\color{blue}7}}}{1}
      {8}{{{\color{blue}8}}}{1}
      {9}{{{\color{blue}9}}}{1}
      {:}{{{\color{red}{:}}}}{1}
      {,}{{{\color{red}{,}}}}{1}
      {\{}{{{\color{black}{\{}}}}{1}
      {\}}{{{\color{black}{\}}}}}{1}
      {[}{{{\color{black}{[}}}}{1}
      {]}{{{\color{black}{]}}}}{1}
      {"}{{{\color{orange}{"}}}}{1},
}

\begin{document}

\title{Bench2Drive-VL: Benchmarks for Closed-Loop Autonomous Driving with Vision-Language Models}

\author{Xiaosong Jia,
Yuqian Shao,
Zhenjie Yang,
Qifeng Li,
Zhiyuan Zhang,
Junchi Yan$^\textsuperscript{\Letter}$,~\IEEEmembership{Senior Member,~IEEE}
\thanks{Corresponding author: Junchi Yan (yanjunchi@sjtu.edu.cn).}
\thanks{Xiaosong Jia are with Institute of Trustworthy Embodied AI, Fudan University and Shanghai Key Laboratory of Multimodal Embodied AI, Shanghai, 200441, China.} 
\thanks{Yuqian Shao, Zhenjie Yang, Qifeng Li, Zhiyuan Zhang, and Junchi Yan are with School of Computer Science \& School of Artificial Intelligence, Shanghai Jiao Tong University, Shanghai, 200240, China.}
}



\maketitle
\IEEEpubidadjcol
\begin{abstract}
With the rise of vision-language models (VLM), their application for autonomous driving (VLM4AD) has gained significant attention. Meanwhile, in autonomous driving, closed-loop evaluation has become widely recognized as a more reliable validation method than open-loop evaluation, as it can evaluate the performance of the model under cumulative errors and out-of-distribution inputs. However, existing VLM4AD benchmarks evaluate the models' scene understanding ability under open-loop, i.e., via static question-answer (QA) dataset. This kind of evaluation fails to assess the VLM's performance under out-of-distribution states rarely appeared in the human collected datasets. To this end, we present Bench2Drive-VL, an extension of Bench2Drive that brings closed-loop evaluation to VLM-based driving, which introduces: (1) DriveCommenter, a closed-loop generator that automatically generates diverse, behavior-grounded question–answer pairs for all driving situations in CARLA—including severe off-route and off-road deviations previously unassessable in simulation. (2) A unified protocol and interface that allows modern VLMs to be directly plugged into the Bench2Drive closed-loop environment to compare with traditional agents. (3) A flexible reasoning and control framework, supporting multi-format visual inputs and configurable graph-based chain-of-thought execution. (4) A complete development ecosystem. Together, these components form a comprehensive closed-loop benchmark for VLM4AD. All codes and annotated datasets are  available at \url{https://github.com/Thinklab-SJTU/Bench2Drive-VL} and \url{https://huggingface.co/datasets/Telkwevr/Bench2Drive-VL-base}.

\end{abstract}


\section{Introduction}\label{sec:intro}

\begin{figure*}[t]
    \centering
    \includegraphics[width=0.90\textwidth]{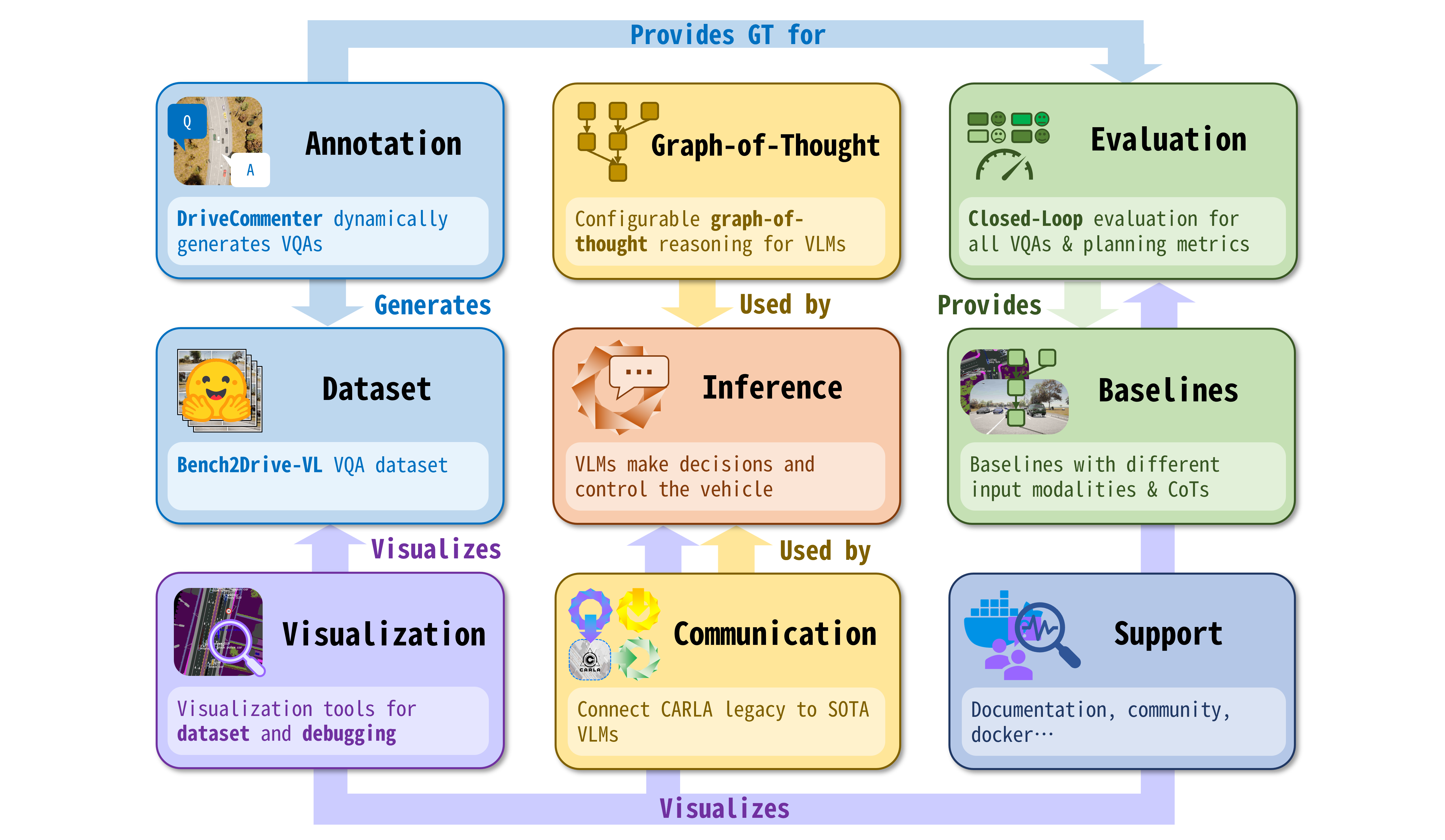}
    \vspace{-4mm}\caption{\textbf{Overview of the Bench2Drive-VL.} 
    (i) The expert model DriveCommenter leverages privileged information to generate ground-truth answers for vision-language questions. (ii) The VLM under evaluation takes annotated sensor data as input, answers the same questions, and controls the ego vehicle. (iii) A VQA evaluator compares the VLM's responses to the ground truth provided by DriveCommenter, while the action module converts the VLM’s natural language instructions into control signals for vehicle actuation.
    }
    \vspace{-6mm}
\end{figure*}

\IEEEPARstart{R}{ecent} advancement of Vision-Language Models (VLMs) has opened new frontiers  for autonomous driving (VLM4AD)~\citep{yang2023survey, LMDrive, DriveWithLLMs, Seff2023motionLM, Jin2023Adapt, tian25DriveVLM, wayve2023lingo1, zhou2025autovla, yang2025drivemoe}. These models jointly interpret visual scenes and natural language instructions, allowing perception, reasoning, and decision-making to be integrated into a unified framework. This integration is essential for the transition from virtual intelligence to the physical world, where driving agents must ground their reasoning in real-time interactive control~\citep{DriveGPT4, DriveMLM, SimLingo, ORION, han2025dmedriver, AlphaDrive, li2025recogdrive}..

Existing benchmarks for VLM4AD~\citep{DriveBench, DrivingVQA, VLADBench, DriveLM} largely focus on open-loop evaluation using static vision-language question-answering (VQA) datasets. Although these datasets offer insights into perception and reasoning, \textbf{they cannot capture the temporal dependencies and cumulative errors that arise during interactive execution}. Since these benchmarks are based on expert distributions, they fail to assess how an agent responds to out-of-distribution states or recovery-critical scenarios induced by its own previous actions. Prior research~\citep{isEgoStatusAllYouNeed, ADMLP, carla_garage_common_mistakes} indicates that such open-loop metrics often overestimate current models' driving proficiency, leaving a significant gap in evaluating closed-loop robustness.

As recently widely acknowledged in the end-to-end autonomous driving community, 
closed-loop evaluation~\citep{NAVSIM, Bench2Drive}  offers a more reliable method for assessing driving capability because it accounts for how an agent’s actions influence future states. However, existing closed-loop frameworks are primarily designed for trajectory-planning systems and \textbf{lack the capacity to evaluate vision-language reasoning dynamically}.  The issue becomes restrictive for reinforcement learning of VLM based AD policy, which require accurate, on-policy ground-truth supervision across the full state distribution, including out-of-distribution scenarios induced by agents' own behaviors. Without such capability, the true performance and learning potential of VLM-based driving agents remain fundamentally underexplored.

This gap presents a fundamental requirements: \textbf{how can we evaluate VLMs' full spectrum of capabilities within a dynamic, interactive environment?} The key technical barrier lies in generating high-quality ground-truth answers to complex driving questions in real time. Manual annotation is labor-intensive and error-prone, while naive LLM/VLM annotation can be slow and unreliable. Rule-based approaches offer higher efficiency, but the combinatorial space of situations is vast. Moreover, answers must be causally consistent with the agent’s behavior, requiring a “know-what-to-do” and “know-why-it-works” expert capable of handling all scenarios.

To this end, we present \textbf{Bench2Drive-VL}, a framework for the closed-loop evaluation of VLM-based driving agents. Based on the widely used  Bench2Drive, Bench2Drive-VL introduces DriveCommenter, an expert agent that automatically generates VQA labels in real time across all simulation states. This allows the system to provide accurate QA labels even for off-road and recovery scenarios that are typically missing from static datasets. Additionally, the framework includes a unified protocol to connect VLMs directly to the CARLA simulator and a graph-of-thought interface for multi-step reasoning. 


The main contributions of this work are summarized as follows. First, we propose \textbf{Bench2Drive-VL}, a framework that unifies multimodal inputs, reasoning chains, and driving actions for the end-to-end closed-loop evaluation of VLM-based agents. Second, we introduce\textbf{ DriveCommenter}, an automated system that generates behavior-consistent VQA labels for all simulation states, facilitating interactive evaluation and reinforcement learning. Third, we implement a flexible reasoning and control module that supports \textbf{graph-based chain-of-thought} execution and heterogeneous perception modalities. Finally, we provide a complete \textbf{development ecosystem}, including visualization tools, debugging utilities, and large-scale annotated datasets to support the benchmarking of reasoning and decision-making capabilities in AD.



\section{Related works}\label{sec:work}

\begin{table*}[!t]
\centering
\caption{
\textbf{Comparison of VLM4AD Benchmarks.}
\label{tab:benchmark-comparison}
}
\resizebox{\textwidth}{!}{%
\begin{tabular}{lcccccccc}
\toprule
\multirow{2}{*}{\textbf{Benchmark}} & \multicolumn{4}{c}{\textbf{Open-loop}} & \multicolumn{4}{c}{\textbf{Closed-loop}} \\
\cmidrule(lr){2-5} \cmidrule(lr){6-9}
& \textbf{Percept.} & \textbf{Predict.} & \textbf{Plan.} & \textbf{Behavior}
& \textbf{Percept.} & \textbf{Predict.} & \textbf{Plan.} & \textbf{Behavior} \\
\midrule
LangAuto\citep{LMDrive} & & & & & & & & CARLA \\
DriveBench\citep{DriveBench} & VQA & VQA & VQA & VQA & & & & \\
DrivingVQA\citep{DrivingVQA} & VQA & VQA & VQA & VQA & & & & \\
DriveLMM-o1\citep{DriveLMM-o1} & VQA & VQA & VQA & VQA & & & & \\
VLADBench\citep{VLADBench} & VQA & VQA & VQA & traj. L2 & & & & \\
\midrule
\rowcolor{gray!10} \textbf{Bench2Drive-VL (Ours)} & VQA & VQA & VQA & VQA
& VQA & VQA & VQA & CARLA\\ 
\bottomrule
\end{tabular}%
}
\vspace{-5mm}
\end{table*}

\subsection{Vision-Language Models for Autonomous Driving}
Recent advances in large vision-language models (VLMs) have demonstrated strong multimodal understanding and reasoning capabilities by jointly modeling visual perception and natural language semantics~\citep{radford2021clip, li2023blip2, alayrac2022flamingo, liu2023llava, openai2024gpt4technicalreport}. These models are increasingly capable of handling complex tasks, such as spatial perception~\citep{Chen2024SpatialVLM, Cheng2024SpatialRGPT, Wu2023Referring}, multi-agent interaction reasoning, and long-horizon decision making. 

Building on these VLM foundations, recent work has extended vision-language reasoning toward embodied settings by incorporating action grounding, giving rise to vision-language-action (VLA) models for robotics and autonomous systems~\citep{PaLM-E, black2024_pi0, black2025_pi0.5, zitkovich-rt2-23a}. 
In such systems, language serves as a flexible interface for perception grounding, high-level planning, and decision explanation, highlighting the importance of carefully evaluating both spatial and reasoning capabilities in interactive driving scenarios.

In autonomous driving (AD), large language models and large vision-language models are appealing for their strong capabilities in visual understanding and reasoning~\citep{hwang2025emma, LMDrive, hu2022ST-P3, DriveGPT4, Pan2024VLP, zhou2025opendrivevla, huang2025drivemm, ORION}. 
These capabilities enable VLMs to provide rich scene analyzing~\citep{Ma2024Dolphins, Cao2024MapLM, Ding2025HiLMD, MTD-GPT}, natural language control~\citep{Park2024VLAAD, cui2024driveasyousay, Cui2024DriveAsYouSpeak, Yang2024humancentric, lübberstedt2025v3lma}, high-level intention prediction~\citep{DriveLM, mao2023gptdriver}, explainable reasoning process~\citep{Wang2025OmniDrive, qian2025agentthink, liu2025xdriver, diao2026DriveRX}, capability of learning from knowledge and experience~\citep{DiLu, wang2024driveanywhere, BaumannN25Enhancing}, which are difficult to achieve with traditional approaches. At the same time, VLMs still exhibit limitations in precise spatial perception and low-level action execution, motivating approaches that decouple VLM-based reasoning from dedicated perception~\citep{OpenEMMA} and control modules~\citep{jiang2024senna, yang2025drivemoe, xu2025humancentricautonomousdrivingfastslow}. All these works have spurred the development of VLM-centric benchmarks and evaluation protocols for autonomous driving, aimed at assessing a full-spectrum of VLMs' capabilities related to driving performance, which we review next.

\vspace{-2.5mm}\subsection{Open-Loop VLM4AD Benchmarks}
Open-loop evaluation remains the dominant approach for assessing vision-language models (VLMs) in autonomous driving due to its scalability and the ease of obtaining structured annotations.  Early datasets such as BDD-X~\citep{BDD-X}, HAD~\citep{HAD}, NuPrompt~\citep{NuPrompt}, LingoQA~\citep{LingoQA}, DRAMA~\citep{DRAMA}, Rank2Tell~\citep{Rank2Tell} etc~\citep{Talk2Car, Ding2024nuinstruct, Chen2025CODA-LM, lu2024lvlmsobtaindriverslicense, godbole2025DRAMA-X, stsbench, jiang2025structuredlabelingenablesfaster, zhou25TUMTrafVideoQA}. typically cover one or more aspects of perception, prediction, and planning through structured question-answering, providing a foundation for evaluating scene understanding in driving.

A more systematic effort is made by DriveLM~\citep{DriveLM}, which introduces a VQA-Graph formulation spanning perception, prediction, planning, and behavior reasoning. By explicitly encoding inter-task dependencies, DriveLM enables the evaluation of cross-task reasoning consistency. 
Building on this, DriveBench~\citep{DriveBench} establishes a GPT-based scoring pipeline that assesses VLM outputs along multiple dimensions. More recently, STRIDE-QA~\citep{Keishi2025strideqa}, NuInteract~\citep{zhao2025nuinteract}, NuScences-SpatialQA~\cite{tian2025NuScenesspatialqa} and WOMD-Reasoning~\citep{Li2025WOMDReasoning} focus on spatiotemporal reasoning in driving scenes, and NuPlanQA~\citep{Park2025nuplanQA} focuses on multi-view driving scene understanding.

More recent benchmarks extend beyond single-turn VQA to incorporate chain-of-thought (CoT) reasoning. Reason2Drive~\citep{Reason2Drive} proposes a reasoning-centric benchmark that decomposes driving understanding into structured chains spanning perception, prediction, and decision making. DrivingVQA~\citep{DrivingVQA} provides natural-language reasoning explanations for each question, whereas DriveLMM-o1~\citep{DriveLMM-o1} goes further by annotating step-by-step CoT traces for all sub-tasks. Although such rich annotations help analyze reasoning quality, they also introduce potential redundancy. On a different axis, VLADBench~\citep{VLADBench} emphasizes scenario diversity and fine-grained categorization. It evaluates VLMs across challenging situations such as occlusion-heavy scenes, rare agent interactions, and complex spatial layouts, offering broader coverage of real-world complexity.


\vspace{-2.5mm}\subsection{Closed-Loop Autonomous Driving Evaluation} Closed-loop evaluation offers a more realistic measurement of autonomous driving capability, as it requires models to interact with the environment and execute actions that influence future states. For VLM-driven agents, LMDrive~\citep{LMDrive} represents the earliest attempt to introduce language-conditioned closed-loop assessment. Its LangAuto benchmark uses the CARLA Town05Long route and requires the model to follow natural-language navigation instructions. Although this framework brings linguistic input into the control loop, its evaluation remains centered around trajectory planning and does not examine whether the VLM can understand the underlying scene, justify its decisions, or reason about causality.

Subsequent works, such as SimLingo~\citep{SimLingo} and ORION~\citep{ORION}, adopt more robust benchmarks including CARLA Leaderboard v2~\citep{CARLALB2} and Bench2Drive~\citep{Bench2Drive}. These frameworks provide diverse routes and corner-case scenarios with well-established safety and efficiency metrics. However, in all these systems, the VLM only produces high-level driving commands or discrete behavior labels, and the evaluation focuses solely on trajectory or maneuver correctness. \textbf{They do not consider multimodal reasoning quality, do not generate VQA ground truth dynamically}.

\vspace{-2.5mm}\subsection{Chain-of-Thought Reasoning in Autonomous Driving} Chain-of-Thought (CoT) prompting was originally proposed as an effective strategy to elicit multi-step reasoning in large language models by explicitly generating intermediate reasoning steps~\citep{wei2022chainofthoughtprompting}. Rather than treating reasoning as a black-box mapping from inputs to outputs, CoT exposes the latent inference process, enabling improved performance on tasks requiring compositional, mathematical, or symbolic reasoning. Subsequent studies have further interpreted CoT as a form of inference-time scaling, where increased computational effort during generation leads to stronger reasoning capabilities~\citep{StepByStep}. Recent research has extended CoT reasoning from pure language models to large vision-language models (VLMs)~\citep{Mitra2024compositionalchainofthoughtprompting, shao2024visualcotadvancingmultimodal}, then further in the growing fields of embodied intelligence~\citep{cosmos_reason1_2025, black2025_pi0.5} and autonomous driving. A growing line of work—such as AlphaDrive~\citep{AlphaDrive}, DriveLMM-o1~\citep{DriveLMM-o1}, CoT-Drive~\citep{liao2025cotdrive}, DriveR1~\citep{DriveR1}, DriveAgent-R1~\citep{DriveAgentR1}, and AdaThinkDrive~\citep{AdaThinkDrive}—demonstrates that incorporating CoT or long-horizon reflective reasoning, combined with reinforcement fine-tuning (RFT), can improve closed-loop driving quality. These models generally build on VLM–based architectures, consume multi-view RGB inputs with short histories, and adopt a multi-stage training paradigm: supervised fine-tuning followed by RL-style policy optimizations. They also explore diverse output parameterizations, ranging from natural-language waypoint descriptions to discretized action tokens or physically grounded motion vocabularies. While promising, these approaches remain evaluated by traditional metrics-either trajectory-centered, such as Bench2Drive~\citep{Bench2Drive}, NAVSIM~\citep{NAVSIM} and nuPlan~\citep{nuplan}, or completely open-loop, like DriveBench~\citep{DriveBench}, VLADBench~\citep{VLADBench}, DrivingVQA~\citep{DrivingVQA} and DriveLMM-o1~\citep{DriveLMM-o1}. They do not assess whether the VLM-based agent is capable of understanding and reasoning properly in evolving environment states.

Altogether, the field still lacks a unified setup that (i) exposes VLMs to interactive closed-loop dynamics, (ii) evaluates both action quality and reasoning ability, and (iii) enables fair comparison with existing closed-loop autonomous driving methods such as Bench2Drive~\citep{Bench2Drive}. Bench2Drive-VL addresses this fundamental gap, and on top of this core capability, it further supports flexible CoT setup for VLMs, provides natural-language annotations over all possible states to facilitate reinforcement learning, and establishes a standardized communication protocol bridging the legacy CARLA simulator with modern large-model architectures—offering researchers a plug-and-play experience.

\section{Methods}

\vspace{-2.5mm}\subsection{Overview}

\begin{figure*}[t]
    \centering
    \includegraphics[width=1.0\textwidth]{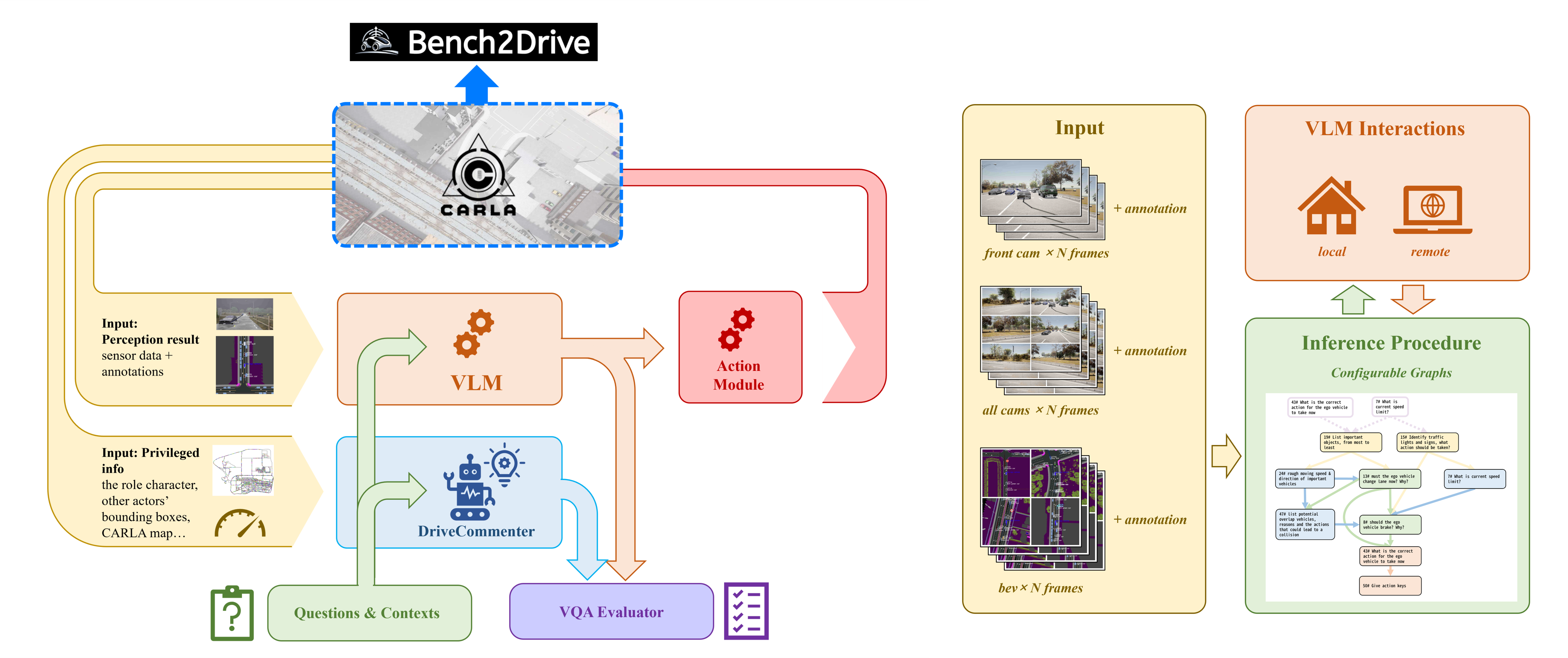}
    \vspace{-5mm}
    \caption{\small{\textbf{Overview of the Bench2Drive-VL closed-loop evaluation framework.}
    The expert model DriveCommenter leverages privileged information to generate ground-truth answers for vision-language questions. 
    The VLM under evaluation takes sensor data as input, answers the same questions, and controls the ego vehicle. 
    A VQA evaluator compares the VLM's responses to the ground truth provided by DriveCommenter, while the action module converts the VLM’s natural language instructions into control signals.}}
    \vspace{-6mm}
    \label{fig:b2d_overview}
\end{figure*}

The\textbf{ Bench2Drive-VL} framework provides a unified closed-loop pipeline (\autoref{fig:b2d_overview}) for evaluating VLM-based driving agents within an interactive simulator. At each intervention interval, the CARLA simulator transmits privileged state information to \textbf{DriveCommenter}, an expert model that generates ground-truth answers for a complex set of vision-language questions based on current environment. Simultaneously, the evaluated VLM policy processes raw sensor inputs to produce its  reasoning and control outputs. To speed up the evaluation process, all  expert answers and VLM predictions are stored and the scoring is conducted offline.  A dedicated VQA evaluator then compares these responses using LLM-based rubric functions, enabling fine-grained assessment of both perception and reasoning quality.
Among these questions, one corresponds to action selection and then an action module converts the VLM’s predicted action from this question into control signals for closed-loop driving. 
Planning quality is subsequently evaluated using Bench2Drive metrics.

Beyond the evaluation pipeline, Bench2Drive-VL provides several subsystems that facilitate dataset construction, model development, and reproducible benchmarking of VLM4AD:


\begin{figure}[h]
    \centering
    \begin{minipage}[b]{0.495\textwidth}
        \centering
        \includegraphics[width=\textwidth]{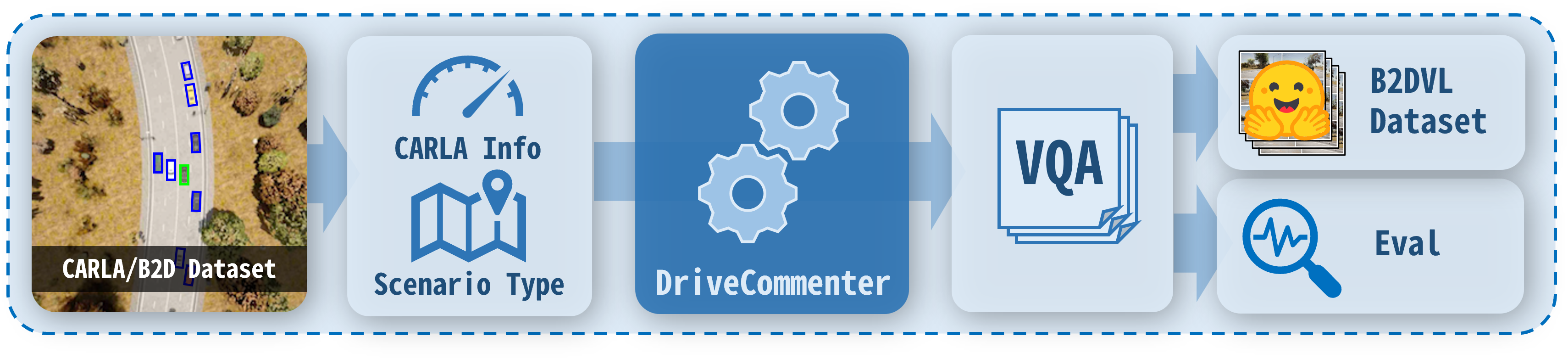}
        \vspace{-8mm}
        \caption{\small{\textbf{Annotation Process of Bench2Drive-VL.}}}
        \label{fig:c_anno}
    \end{minipage}
    \hfill
    \begin{minipage}[b]{0.495\textwidth}
        \centering
        \includegraphics[width=\textwidth]{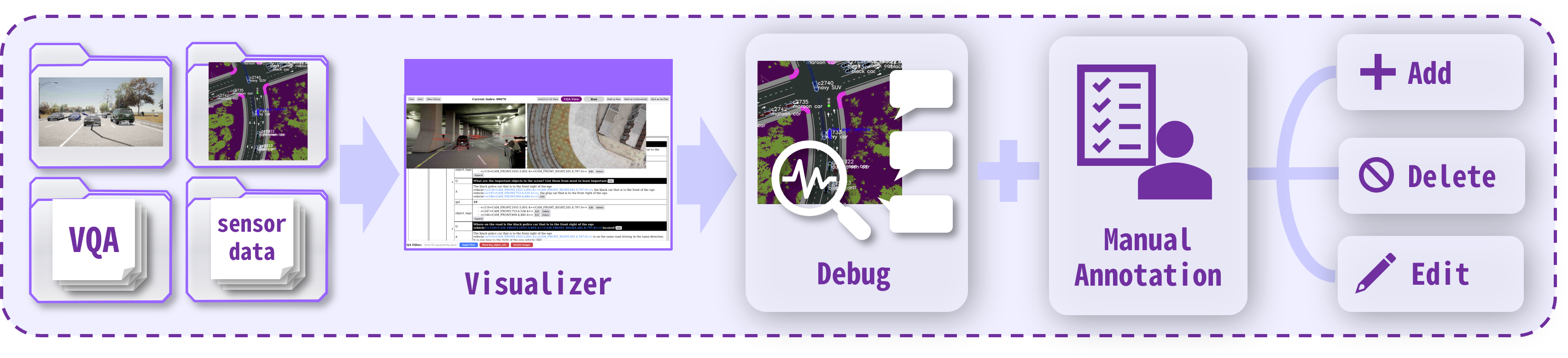}
        \vspace{-8mm}
        \caption{\small{\textbf{Visualization Tools of Bench2Drive-VL.}}}
        \vspace{-6mm}
        \label{fig:c_vis}
    \end{minipage}
\end{figure}

\noindent\textbf{Annotation and Dataset Construction} (\autoref{fig:c_anno}, \autoref{fig:c_vis})  
provide automatic and human-in-the-loop tools for building training data and inspecting closed-loop behavior.  
DriveCommenter generates structured questions and natural-language answers about vehicles, pedestrians, traffic signals, lane geometry, and weather conditions.  
A visualization and annotation GUI supports correction of mislabeled data, addition of new QA pairs, and debugging of the VLM’s reasoning graph.

\begin{figure}[h]
    \centering
    \begin{minipage}[b]{0.495\textwidth}
        \centering
        \includegraphics[width=\textwidth]{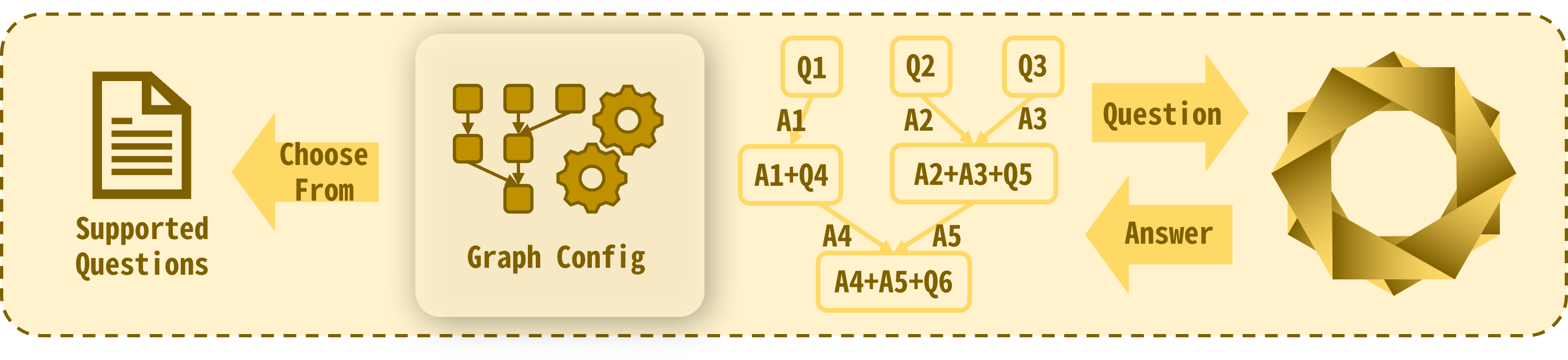}
        \vspace{-8mm}
        \caption{\small{\textbf{Graph-of-Thought of Bench2Drive-VL.}}}
        \label{fig:c_graph}
    \end{minipage}
    \hfill
    \begin{minipage}[b]{0.495\textwidth}
        \centering
        \includegraphics[width=\textwidth]{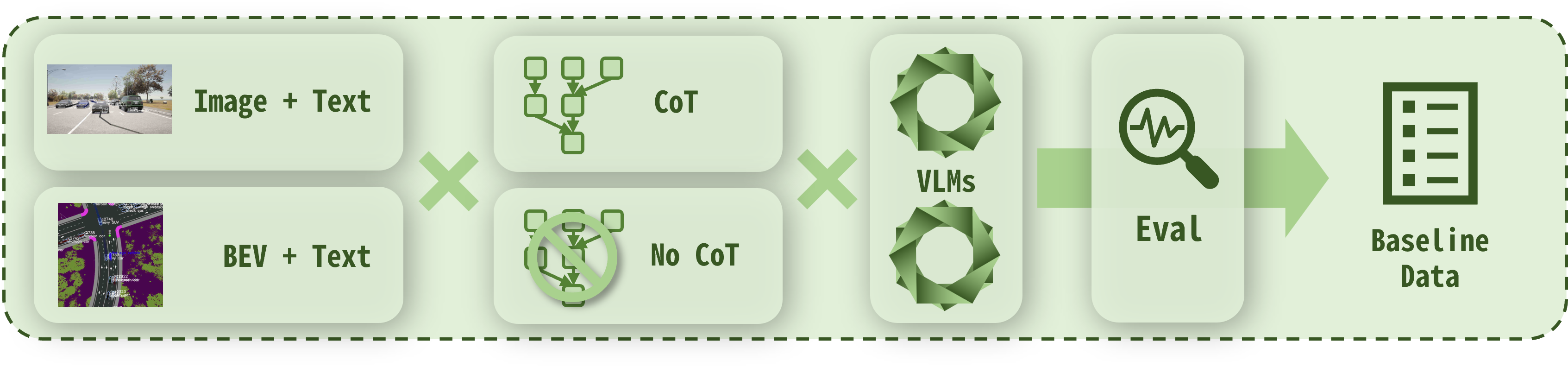}
        \vspace{-8mm}
        \caption{\small{\textbf{Baselines of Bench2Drive-VL.}}}
        \label{fig:c_baseline}
    \end{minipage}
    \vspace{-8mm}
\end{figure}

\noindent\textbf{Reasoning Configuration and Baseline Models} (\autoref{fig:c_graph}, \autoref{fig:c_baseline})  
support flexible experimentation with different VLM reasoning paradigms.  
The Graph-of-Thought configuration system allows users to define customized inheritance structures and multi-step reasoning chains.
A suite of image-, BEV-, and text-based VLM baselines is included as starting points for model comparison, as reported in \autoref{tab:model_comparison_behavior} and \autoref{sec:experiments}.

\begin{figure}[h]
    \centering
    \begin{minipage}[b]{0.495\textwidth}
        \centering
        \includegraphics[width=\textwidth]{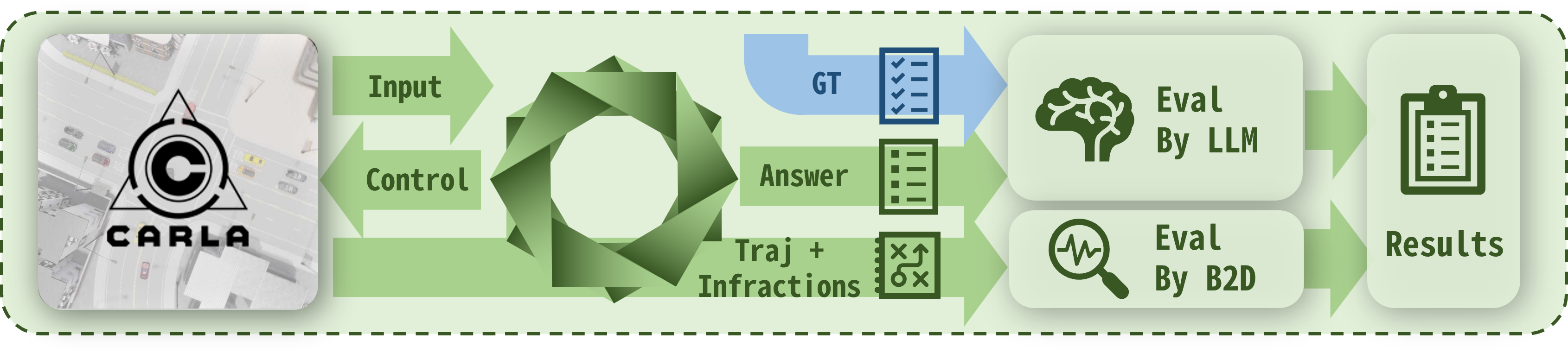}
        \vspace{-8mm}
        \caption{\small{\textbf{Closed-Loop Evaluation Process of Bench2Drive-VL.}}}
        \label{fig:c_eval}
    \end{minipage}
    \hfill
    \begin{minipage}[b]{0.495\textwidth}
        \centering
        \includegraphics[width=\textwidth]{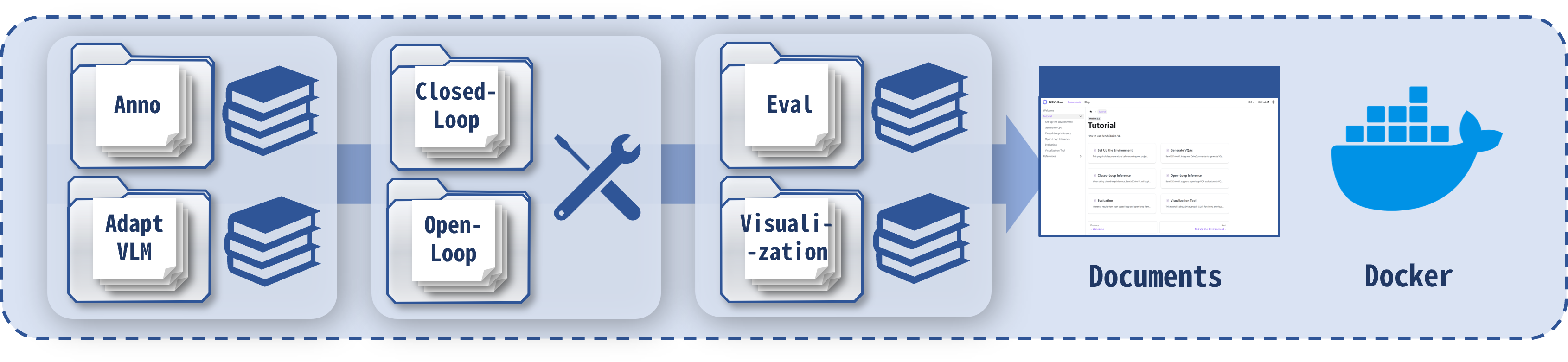}
        \vspace{-8mm}
        \caption{\small{\textbf{Utilities Provided by Bench2Drive-VL.}}}
        \vspace{-4mm}
        \label{fig:c_doc}
    \end{minipage}
\end{figure}

\noindent\textbf{Closed-Loop Runtime and System Utilities} (\autoref{fig:c_eval}, \autoref{fig:c_doc})  
provide the executable layer for integrating CARLA with modern large-model architectures.  
Bench2Drive-VL supports distributed deployment where the simulator and VLM run on different machines through a web-based communication protocol.  
Comprehensive documentation and Docker environments address the well-known challenges of configuring CARLA for closed-loop evaluation, ensuring reproducibility across different platforms.

\vspace{-2.5mm}\subsection{DriveCommenter} 

\begin{figure*}[t]
    \centering
    \includegraphics[width=0.85\textwidth]{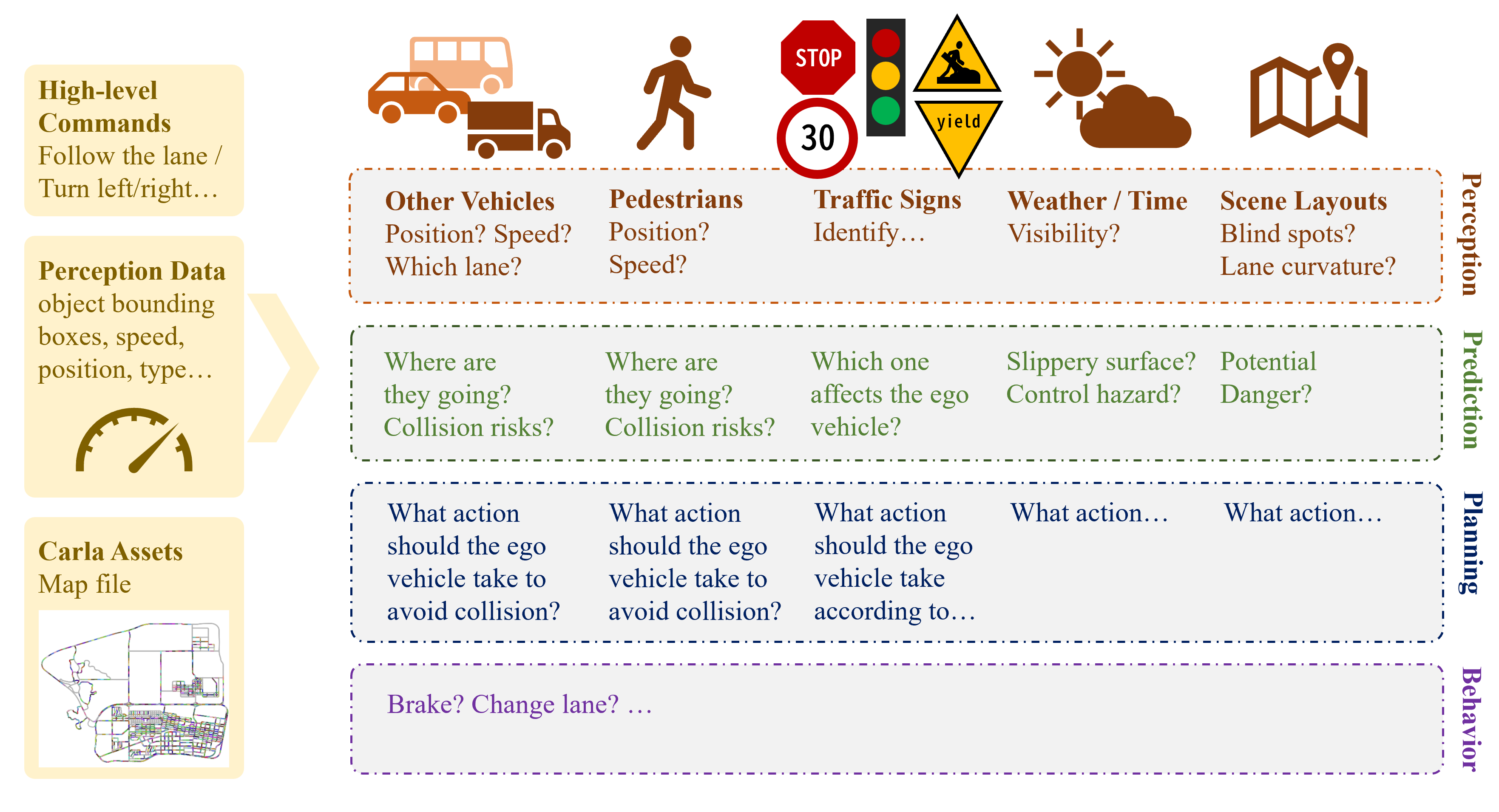}
    \vspace{-3mm}
    \caption{\small{\textbf{Overview of DriveCommenter.} Our VQAs cover all tasks (perception, prediction, planning, behaviour) related to all categories (vehicles, pedestrians, traffic signs, weather and scene layouts).}}
    \vspace{-7mm}
    \label{fig:dc}
\end{figure*}

\noindent\textbf{Key Distinctions from previous pipelines}. DriveCommenter is not the first tool to obtain vision–language datasets based on CARLA. The VQA-generation script included DriveLM's code \cite{DriveLM} uses the decision information produced by PDM-Lite\cite{beisswenger2024pdmlite} together with sensor data collected in CARLA to construct the DriveLM-CARLA dataset. PDM-Lite, inspired by PDM-Closed~\cite{dauner23aPDM-Closed} for nuPlan~\cite{nuplan}, is a rule-based planner that uses the Intelligent Driver Model (IDM)~\cite{treiber2000idm} to determine target speeds based on the ego vehicke's surroundings. Most questions that this script can annotate are limited to perception and final planning, resulting in relatively narrow coverage. The goal of DriveCommenter is to directly construct VQA datasets from sensor information, and therefore its decision-making must be performed within its own pipeline rather than relying on the DriveLM-CARLA-style generation script.

\noindent\textbf{Diverse behavior generation for robust QA.} Moreover, the decisions made by PDM-Lite tend to be single mode, whereas human decisions can often be multi-mode. For example, when lane-changing to bypass an obstacle, PDM-Lite first generates the closest waypoint that envelopes the obstacle and then plans based on that route. This means that a vehicle controlled by PDM-Lite will inevitably approach the obstacle as closely as possible before waiting for an opportunity to change lanes. In practice, however, lane-change considerations can begin as soon as the obstacle is observed. Thus, the key advantages of DriveCommenter as an expert model for diverse behavior generation are as follows:
\begin{enumerate}[leftmargin=10pt, topsep=0pt, itemsep=1pt, partopsep=1pt, parsep=1pt]
\item No reliance on expert-model decision information.
\item Optimizing the question structure by merging simple object-wise questions into holistic ones, reducing answer length.
\item Incorporating objects outside the front-view camera range, which is particularly important for interactive behaviors like overtaking and yielding to emergency vehicles.
\item Adding a large number of questions related to driving-environment perception, high-level driving decisions, and additional traffic-sign categories.
\item Adding handling of out-of-distribution scenarios on non-expert trajectories to ensure that the system can perform the tasks required in Bench2Drive-VL.
\end{enumerate}

Comaprision of questions supported by DriveCommenter and DriveLM-CARLA's annotation script are  in \autoref{tab:drivecommenter_qas}. 

DriveCommenter is capable of generating visual question-answer pairs (VQAs) from both  static datasets and running CARLA simulations. Similar to Bench2Drive dataset, all annotated VQAs are organized into JSON files named by frame index, grouped under folders named after the scenario. Each question is assigned a unique ID.









\noindent\textbf{Annotating Static Datasets}. When annotating datasets derived from CARLA, such as the Bench2Drive dataset\citep{Bench2Drive}, DriveCommenter takes a wide range of perception data as input, including the position, velocity, bounding boxes, and type descriptions of all actors in the current scene, as well as the high-level commands assigned to the ego vehicle and environmental conditions such as weather. In addition, the corresponding map file of the scene is required to obtain information such as lane topology.

\noindent\textbf{Connecting to CARLA Simulation}. When connected to a running CARLA, DriveCommenter  takes almost the same set of data as during offline annotation. The only exception is the treatment of special role actors: in certain CARLA Leaderboard v2 scenarios, special vehicles are generated for dynamic interactions, such as vehicles that must be overtaken, red-light violators that require emergency avoidance, etc. DriveCommenter handles these special actors with special processing logic. In offline expert-trajectory datasets, their identities can be inferred via relaxed rule-based filters due to the determinism of routes. While in the CARLA runtime where ego trajectories may vary significantly, DriveCommenter could directly query the scenario configuration information to identify these special actors and react accordingly.

\begin{figure}[t]
    \centering
    \includegraphics[width=1.0\linewidth]{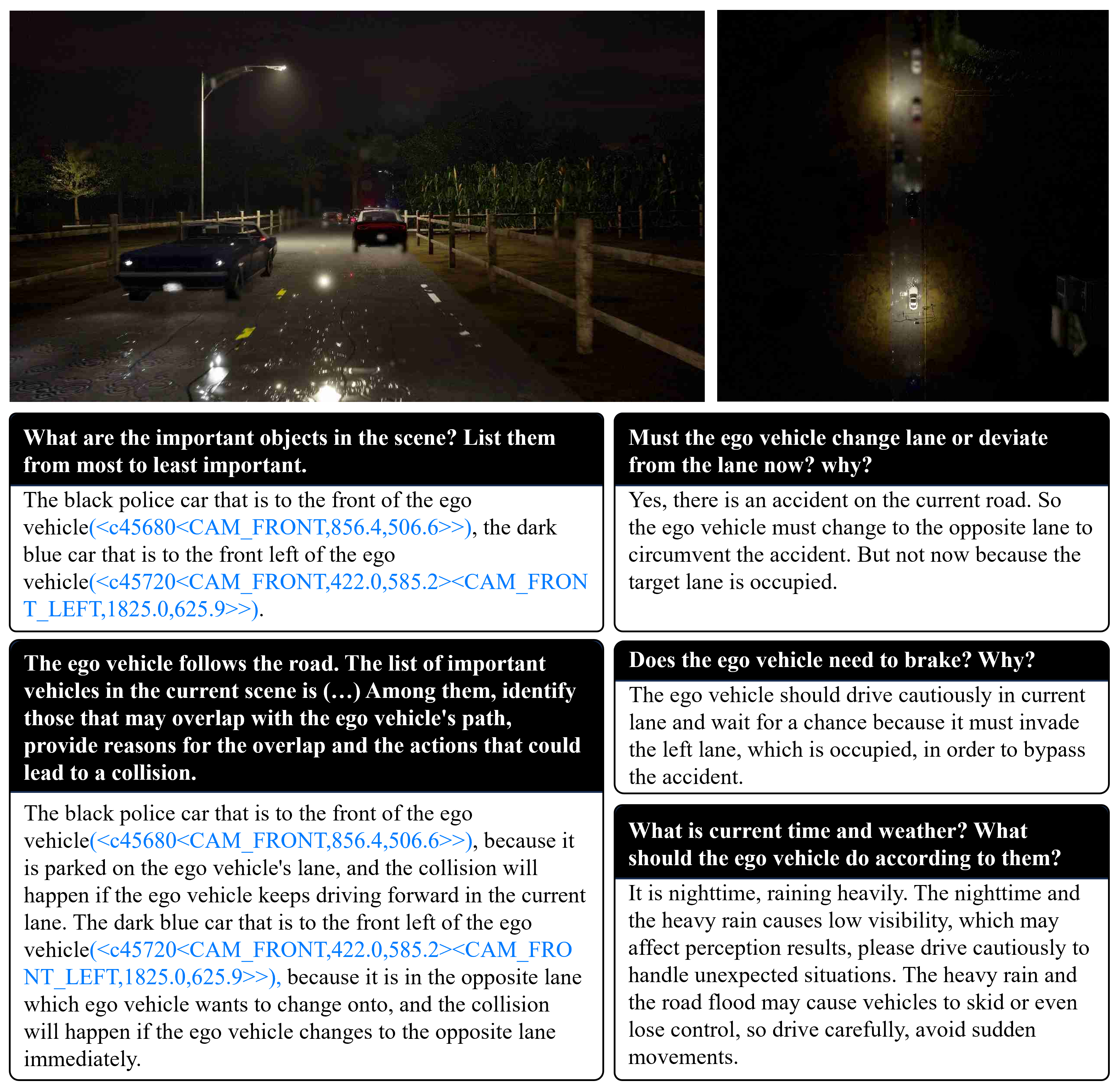}
    \vspace{-2mm}
    \caption{\small{\textbf{Example of A Frame Annotated by DriveCommenter.} Only few questions are shown for space.}}
    \vspace{-8mm}
    \label{fig:accident_frame}
\end{figure}

\definecolor{perceptioncolor}{RGB}{220, 245, 255}
\definecolor{predictioncolor}{RGB}{255, 230, 230}
\definecolor{planningcolor}{RGB}{230, 255, 230}
\definecolor{behaviorcolor}{RGB}{255, 255, 200}

\begin{table*}[htbp]
\centering
\caption{Comparison of questions supported by DriveCommenter and DriveLM-CARLA's automatic annotation method.}
\resizebox{\textwidth}{!}{%
\begin{tabular}{|c|p{12cm}|c|c|}
\hline
\textbf{QID} & \textbf{Question} &
\shortstack{\textbf{DriveLM}\\\textbf{-CARLA}} &
\shortstack{\textbf{Drive}\\\textbf{Commenter}} \\
\hline
\multicolumn{4}{|l|}{\cellcolor{perceptioncolor}\textbf{Perception}} \\
\hline
1 & How many pedestrians are there? & \checkmark & \checkmark \\
2 & Is the ego vehicle affected by a stop sign? & \checkmark & \checkmark \\
3 & Is the ego vehicle affected by a speed limit sign? &  & \checkmark \\
4 & List the traffic signs affecting the ego vehicle in the current scenario. &  & \checkmark \\
5 & Is the ego vehicle affected by a traffic light? & \checkmark & \checkmark \\
6 & What is the state of the traffic light? & \checkmark & \checkmark \\
11 & Is there an obstacle on the current road? & \checkmark & \checkmark \\
18 & What are the important objects in the scene? & \checkmark & \checkmark \\
19 & What are the important objects in the scene? List them from most to least important. &  & \checkmark \\
20 & Where on the road is \{vehicle\_description\} located? & \checkmark & \checkmark \\
26 & What are the important vehicles and where are they on road? & \checkmark & \checkmark \\
27 & The important vehicles are ..., List their locations on road. &  & \checkmark \\
30 & From which side are other vehicles allowed to change lanes into the ego lane? & \checkmark & \checkmark \\
31 & In which direction is the ego car allowed to change lanes? & \checkmark & \checkmark \\
32 & What lane marking is on the \{name\} side of the ego car? & \checkmark & \checkmark \\
33 & On which lane is the ego vehicle (left most lane of the lanes going in the same direction is indicated with 0)? & \checkmark & \checkmark \\
34 & How many lanes are there in the \{name\} direction \{to\_or\_as\} the ego car? & \checkmark & \checkmark \\
35 & Is the ego vehicle at a junction? & \checkmark & \checkmark \\
36 & The ego vehicle wants to \{command\_description\}. Which lanes are important to watch out for? & \checkmark & \checkmark \\
37 & What is current time and weather? &  & \checkmark \\
40 & Apart from vehicles on the road, visible pedestrians and the weather, what other factors in the current scenario could pose potential hazards? &  & \checkmark \\
44 & Describe the current lane's direction. &  & \checkmark \\
\hline
\multicolumn{4}{|l|}{\cellcolor{predictioncolor}\textbf{Prediction}} \\
\hline
16 & What is the moving status of \{other\_vehicle\_location\_description\}? & \checkmark & \checkmark \\
17 & Where is \{other\_vehicle\_location\_description\} going? & \checkmark & \checkmark \\
21 & What is the rough moving speed and moving direction of \{vehicle\_description\}? &  & \checkmark \\
22 & What is the exact moving speed and moving direction of \{vehicle\_description\}? &  & \checkmark \\
23 & The ego vehicle \{command\_str\}. Is \{vehicle\_location\_description\} potentially crossing the path of the ego vehicle? If so, why? &  & \checkmark \\
24 & The important vehicles are ..., What is the rough moving speed and moving direction of them? &  & \checkmark \\
25 & The important vehicles are ..., What is the exact moving speed and moving direction of them? &  & \checkmark \\
28 & The important vehicles are ..., Identify potential overlap vehicles and give reasons. &  & \checkmark \\
29 & The important vehicles are ..., List potential overlap vehicles. &  & \checkmark \\
38 & What is current time and weather? What hazards might it bring? &  & \checkmark \\
39 & What is current time and weather? What should the ego vehicle do according to them? &  & \checkmark \\
46 & The important vehicles are ..., List potential overlap vehicles and the actions that could lead to a collision. &  & \checkmark \\
47 & The important vehicles are ..., List potential overlap vehicles, overlap reasons and the actions that could lead to a collision. &  & \checkmark \\
48 & The ego vehicle \{command\_str\}. Is \{vehicle\_description\_with\_location\} potentially crossing the path of the ego vehicle? If so, why? And what action can lead to a collision? &  & \checkmark \\
49 & The ego vehicle \{command\_str\}. Is \{vehicle\_description\_with\_location\} potentially crossing the path of the ego vehicle? If so, what action can lead to a collision? &  & \checkmark \\
\hline
\multicolumn{4}{|l|}{\cellcolor{planningcolor}\textbf{Planning}} \\
\hline
7 & What is the current speed limit? &  & \checkmark \\
8 & Does the ego vehicle need to brake? Why? & \checkmark & \checkmark \\
9 & What should the ego vehicle do based on the \{actor\_type\}? & \checkmark & \checkmark \\
10 & Does the ego vehicle need to change lanes or deviate from the lane center due to an upcoming obstruction? & \checkmark & \checkmark \\
12 & Does the ego vehicle need to change lanes or deviate from the lane for reasons other than the upcoming obstruction? Why? &  & \checkmark \\
13 & Must the ego vehicle change lane or deviate from the lane now? Why? &  & \checkmark \\
14 & The list of traffic lights and signs affecting the ego vehicle in current scene is: \{sign\_list\_str\}. Based on these traffic signs, what actions should the ego vehicle take respectively? &  & \checkmark \\
15 & Identify all traffic lights and signs affecting the ego vehicle in current scene. Based on these traffic signs, what actions should the ego vehicle take respectively? &  & \checkmark \\
\hline
\multicolumn{4}{|l|}{\cellcolor{behaviorcolor}\textbf{Behavior}} \\
\hline
42 & Predict the ego vehicle's future waypoint... &  & \checkmark \\
43 & What is the correct action for the ego vehicle to take now? &  & \checkmark \\
50 & Provide the appropriate behavior for the ego vehicle, FOLLOW\_LANE, CHANGE\_LANE\_LEFT, CHANGE\_LANE\_RIGHT, GO\_STRAIGHT, TURN\_LEFT, or TURN\_RIGHT and the Speed key, which can be KEEP, ACCELERATE, DECELERATE, or STOP. &  & \checkmark \\
\hline
\end{tabular}%
} %
\label{tab:drivecommenter_qas}
\end{table*}

\noindent\textbf{Diverse and Accurate VQA Label Generation}. DriveCommenter poses questions about virtually every object related to driving and decision-making in the scene, including vehicles, slow-moving actors such as bicycles and pedestrians, traffic signs, the curvature of the ego lane, and more. While the standard CARLA evaluation benchmark typically only concerns stop signs and traffic lights, DriveCommenter expands the annotation space to include speed-limit signs, yield signs, construction warnings, and other traffic-relevant indicators to better match real-world driving requirements. Speed-limit signs, in particular, require contextual understanding: the ego vehicle must remember speed limits that are no longer visible but remain in effect.

DriveCommenter also accounts for the impact of weather and illumination conditions, such as darkness, fog, and rain reducing visibility, or puddles and flooding increasing braking risks. In specific tunnel scenarios in CARLA, questions related to weather or time of day are answered with:
\textit{``It is impossible to infer the current time and weather from visual information, because the ego vehicle is currently inside a tunnel.''}

In some scenarios, pedestrians or cyclists may suddenly emerge from behind parked vehicles or other occluding structures. DriveCommenter incorporates such “blind-spot” risks into safety-related questions by explicitly noting limited visibility or partial occlusions when appropriate.

DriveCommenter generates high-level decision-making questions as well. For example, Question~43 requires the ego vehicle's current driving decision to be described in natural language, whereas Question~50 requests a key–value selection specifying the driving direction and speed. The direction options include
\texttt{[FOLLOW\_LANE, CHANGE\_LANE\_LEFT, CHANGE\_LANE\_RIGHT, GO\_STRAIGHT, TURN\_LEFT, TURN\_RIGHT, DEVIATE\_LEFT, DEVIATE\_RIGHT]}.
The accompanying speed choices include
\texttt{[KEEP, ACCELERATE, DECELERATE, STOP]}.
Among these, \texttt{FOLLOW\_LANE}, \texttt{CHANGE\_LANE\_LEFT}, \texttt{CHANGE\_LANE\_RIGHT}, \texttt{DEVIATE\_LEFT}, and \texttt{DEVIATE\_RIGHT} are used when the ego vehicle is not at an intersection. The two \texttt{DEVIATE} actions represent lateral deviations within the lane; they are needed, for instance, in the \texttt{InvadingTurn} scenario where the ego vehicle must shift laterally to avoid an oncoming vehicle that partially invades its lane. The intersection-related actions \texttt{GO\_STRAIGHT}, \texttt{TURN\_LEFT}, and \texttt{TURN\_RIGHT} describe the ego vehicle's intended movement at junctions.

\begin{figure*}[t]
    \centering
    \includegraphics[width=0.90\textwidth]{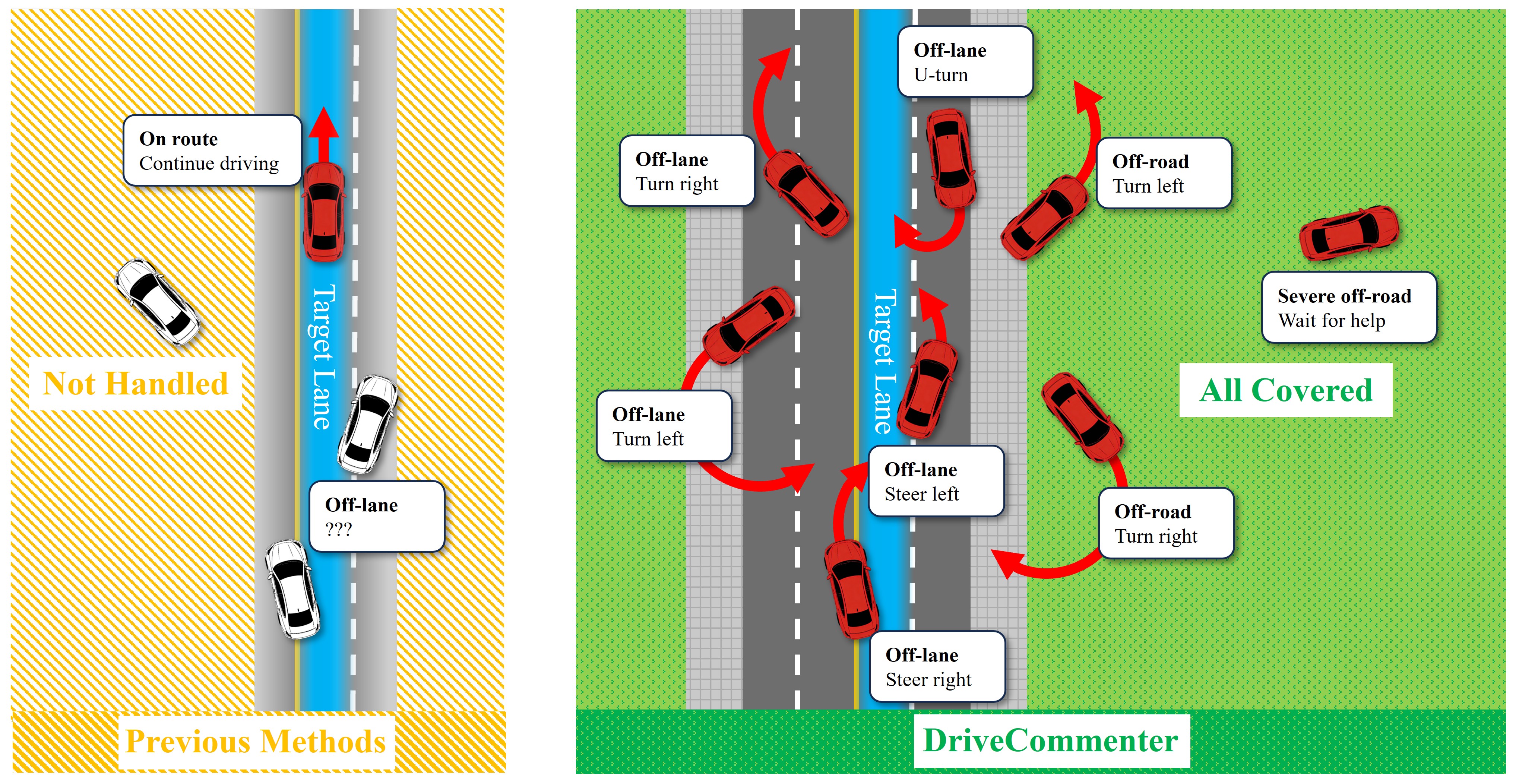}
    \caption{\textbf{Comparison between Existing VLM4AD Annotation Methods and DriveCommenter under Out-of-Distribution (OOD) Conditions.} 
    Existing methods focus on states that lie close to the expert trajectory, leaving a large portion of the simulator state space underexplored. In contrast, DriveCommenter is capable of covering and annotating a much broader set of OOD states.
    }
    \vspace{-7mm}
    \label{fig:overview-ood}
\end{figure*}

\noindent\textbf{Out-of-Distribution (OOD) Handling}.
To support reinforcement learning and guarantee stable supervision in closed-loop evaluation, DriveCommenter must generate valid control actions and coherent natural-language feedback for \emph{all} simulator states, including abnormal or rare ones, as in ~\autoref{fig:overview-ood}. Without explicit OOD handling, trajectories may terminate prematurely, some states may become unlabeled, and large regions of the state space may remain semantically undefined. Therefore, DriveCommenter incorporates a principled OOD module to ensure that every reachable state (on or off road, within or outside the correct lane) receives consistent corrective actions and interpretable explanations.

Road-OOD refers to situations where the ego vehicle leaves the drivable surface. DriveCommenter distinguishes two cases. 

\begin{enumerate}
    \item \textbf{Unrecoverable Road-OOD}: When no valid nearest waypoint can be found, the vehicle is considered completely off road and unable to recover autonomously. The system issues an emergency stop and produces an explanation indicating that the vehicle has left the roadway and must wait for assistance.
    \item \textbf{Recoverable Road-OOD}: When a nearest waypoint exists, DriveCommenter computes lateral deviation direction and commands steering toward the road centerline. The generated text explicitly states that the vehicle has gone off the road and should steer left or right.
\end{enumerate}

Lane-OOD arises when the vehicle remains on the road but deviates from the intended lane. There are three Cases. 

\begin{enumerate}
    \item \textbf{Orientation Misalignment}: If the vehicle's heading diverges from the lane forward vector, DriveCommenter performs progressive alignment based on the deviation angle. Mild misalignment triggers small steering corrections, whereas severe or near-reverse orientations lead to strong steering or even a controlled U-turn. 
    \item \textbf{Lateral Lane Deviation}: When the vehicle is on the correct road but in the wrong lane, DriveCommenter identifies the nearest point on the target lane and determines the required lane-change direction. Safety constraints such as side-lane clearance, obstacle-circumvention status, and emergency-vehicle yielding may delay the maneuver. The language output clearly states both the intended correction (e.g., returning to the left or right lane) and the reason for postponement if applicable.
    \item \textbf{Junction Adaptation}: When the vehicle enters a junction, lane boundaries become less rigid. DriveCommenter therefore relaxes lane-level enforcement and adjusts phrasing of its instructions to turn more naturally.
\end{enumerate}


\vspace{-2.5mm}\subsection{ Diverse Input Configurations for VLM Polices} 

\begin{figure*}[t]
    \centering
    \includegraphics[width=1.0\textwidth]{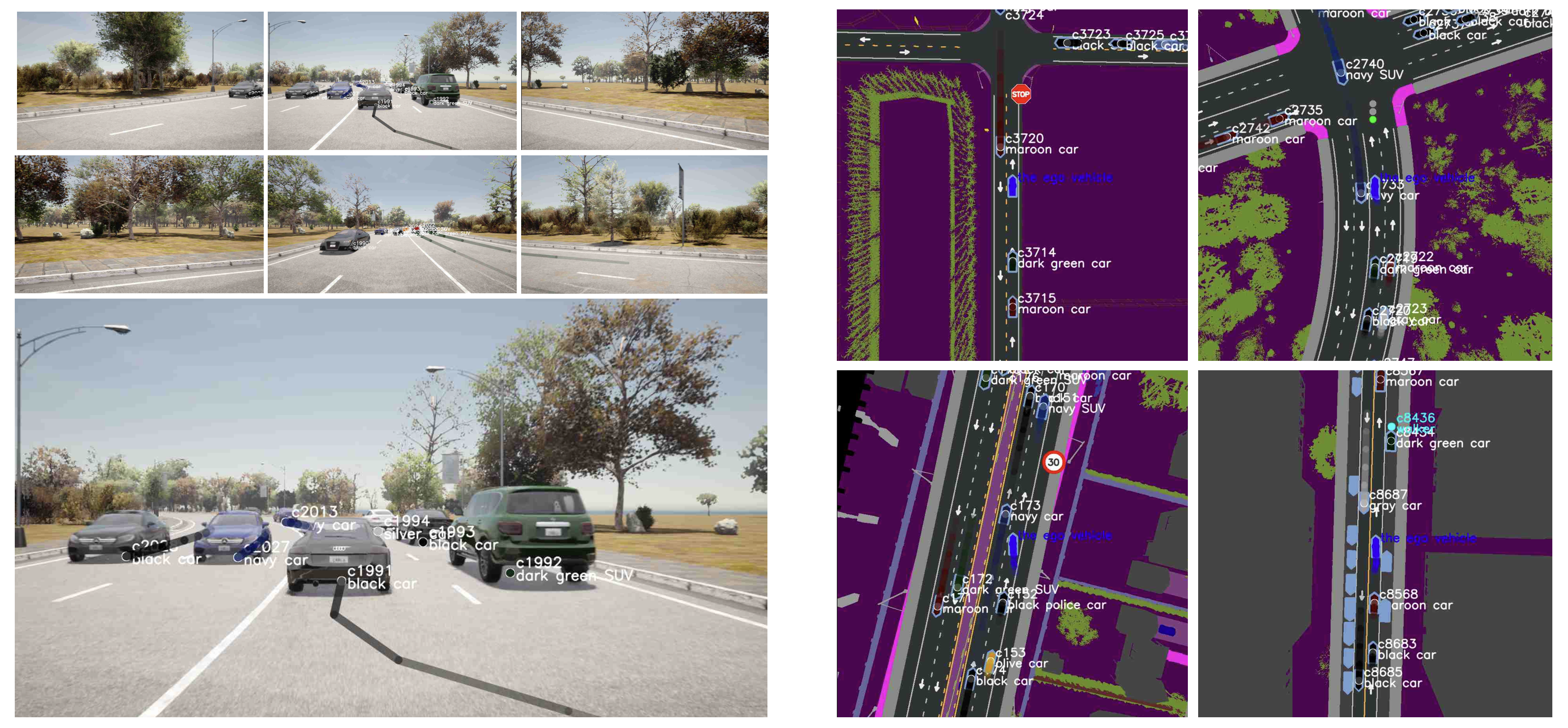}
    \vspace{-4mm}
    \caption{\small{\textbf{Visual Inputs Provided by Bench2Drive-VL.} Annotated RGB Images (left) and BEV Maps (right).}}
    \vspace{-6mm}
    \label{fig:inputs}
\end{figure*}

Bench2Drive-VL provides three kinds of inputs (RGB images, BEV maps, and text) for the evaluated VLMs. 

\noindent\textbf{RGB Input.} For RGB camera layouts, Bench2Drive-VL follows the widely used nuScenes setup. On top of raw images, annotations could be added to help VLMs reference specific objects in VQAs and assist in later evaluation, where a ID and a brief natural language description are labeled in white text, as in~\autoref{fig:inputs} (Left). The massive stationary parked vehicles are not labeled to avoid overlapping labels. The past trajectories of objects could be also annotated using segmented polylines in their respective colors, with each segment anchored at a past position recorded every 0.5 seconds, allowing a single image to convey motion history. It also supports specifying history length and choosing between single-view and multi-view modes. Users can set a historical length to pass images from previous frames to the VLM. 

\noindent\textbf{BEV Input.}  Bench2Drive-VL could also provides BEV map as input, as in ~\autoref{fig:inputs} (Right). The rendering approach is inspired by the \texttt{no\_rendering\_mode} example in CARLA but re-implemented using OpenCV for efficiency.  
The BEV input also supports configuration of historical length.

To generate the BEV image, the renderer first uses a semantic camera mounted above the ego vehicle to get a top-down semantic view as the base layer. Then, using CARLA's map topology and lane information, the road elements (lane markings, directions, sidewalks, etc.) are drawn. Vehicles in the scene are rendered using their own body color for the fill and light blue for the border. Each vehicle is displayed as an arrow to indicate its heading. Just like in the RGB input, the vehicle's description and ID are labeled in white text. Pedestrians are  marked using cyan-blue outlines and labels. To avoid confusion with traffic lights, the ego vehicle is labeled in blue with a "the ego vehicle" tag next to it.

Traffic signs like traffic lights, stop signs, and speed limits are also annotated. They are drawn as icons and projected to their actual positions. Stop signs placed on the road surface at intersections are treated the same as normal upright stop signs. Other traffic signs like yield signs, construction warnings, and construction cones, are marked with yellow dots and labeled accordingly. To reduce confusion, only traffic signs affecting the ego vehicle are shown; others are ignored.

\noindent\textbf{Text Inputs.} Bench2Drive-VL provides essential perception cues in the form of textual prompts. For each vehicle (including bicycles) and pedestrian, the location and velocity are summarized in language. Only relevant objects are included, where those with too few LiDAR points (threshold usually 3), too far from the ego vehicle (typically 50\,m), or too far vertically (usually 30\,m) are excluded.


\vspace{-2.5mm}\subsection{Online Inference of VLM with CARLA Simulator}

Bench2Drive-VL performs VLM inference by raising a customizable set of questions defined in a graph structure. This design allows for flexible chain-of-thought reasoning, where the answer to one question can inform the context of others. Further, the framework decouples the running of CARLA and the VLM via a web-based interface, allowing them to run on separate machines and simplifying deployment for speed up.

\noindent\textbf{Graph-based Reasoning.} The inference module supports highly configurable graph-based reasoning. Users can define, through a configuration file, the set of questions to be answered (\texttt{NODE}), the dependency relationships among them (\texttt{EDGE}), the context inheritance from previous frames (\texttt{INHERIT}), and any nodes for which ground truth should be used (\texttt{USE\_GT}). Before inference, all nodes are topologically sorted to respect dependency ordering. For each question, its immediate predecessors and, if applicable, inherited nodes from previous frames are included in the context. Nodes in \texttt{USE\_GT} directly contribute their ground-truth answers rather than invoking VLM reasoning. This structure allows users to control the reasoning flow, emulate chain-of-thought strategies, and accumulate context across frames when needed. An example configuration is shown below:

\begin{lstlisting}[language=json]
"CHAIN": {
    "NODE": [19, 15, 7, 24, 13, 47, 8, 43, 50],
    "EDGE": {
        "19": [24, 13, 8],
        "15": [7, 8],
        "7": [8],
        "24": [13, 47],
        "13": [47, 8, 43],
        "47": [8],
        "8": [43],
        "43": [50],
        "50": []
    },
    "INHERIT": {
        "19": [43, 7],
        "15": [7]
    },
    "USE_GT": [24]
}
\end{lstlisting}

\begin{figure}[h]
    \centering
    \includegraphics[width=1.0\linewidth]{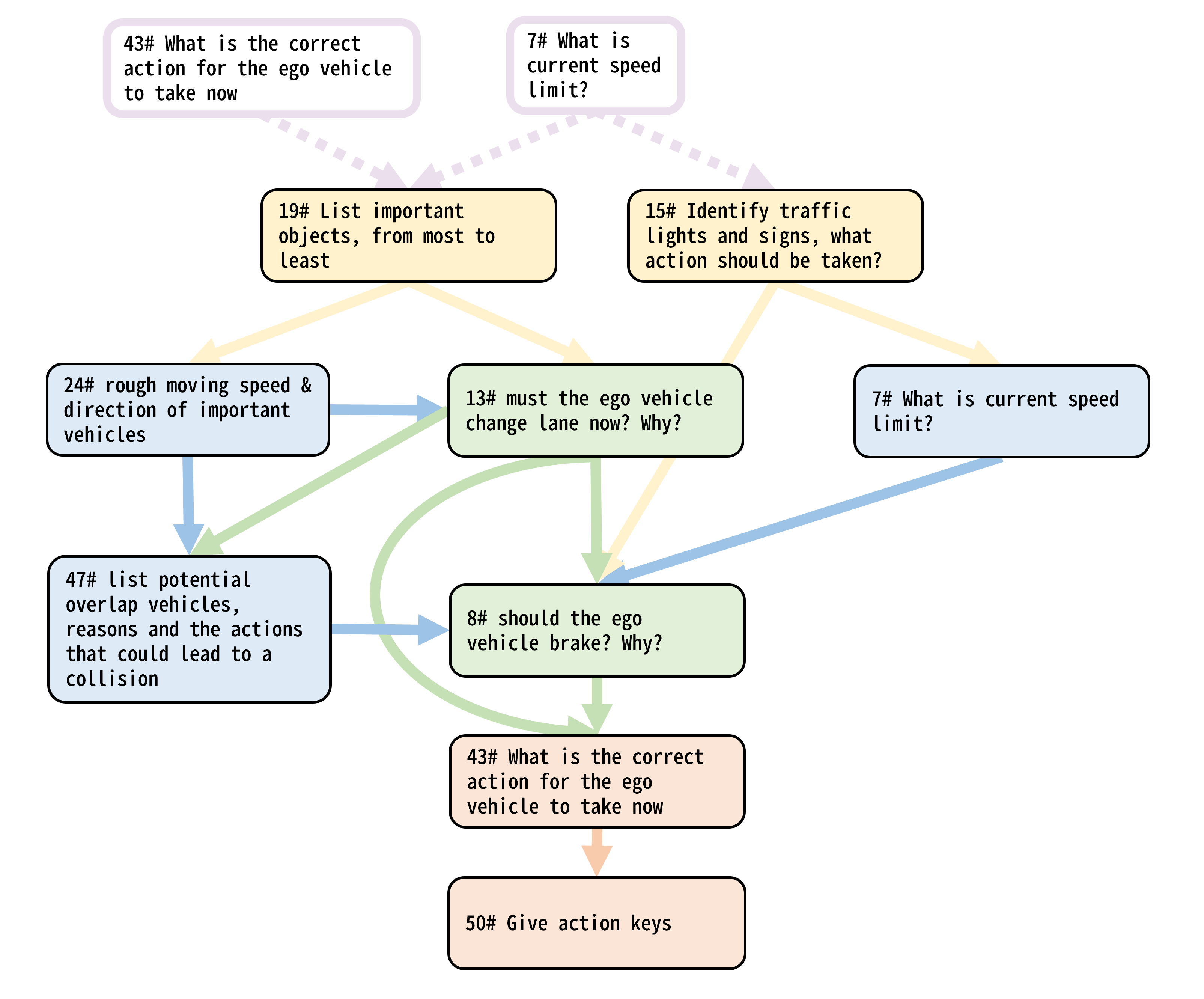}
    \vspace{-4mm}
    \caption{\textbf{The Corresponding Chain-of-Graph  Reasoning Flow.}}
    \vspace{-3mm}
    \label{fig:graph}
\end{figure}

For the example configuration file above, a valid single-frame inference sequence (\autoref{fig:graph}) proceeds as follows:
\begin{enumerate}[leftmargin=10pt, topsep=0pt, itemsep=1pt, partopsep=1pt, parsep=1pt]
    \item Question 19 is answered first, using the answers to Questions 43 and 7 from the previous frame as context.
    \item Question 24 is skipped for VLM inference because it uses the ground truth (\texttt{USE\_GT}).
    \item Question 13 is answered next, using the answers to Questions 19 and 24 from the current frame as context, with Question 24’s answer provided by ground truth.
    \item Question 47 is answered using the answers to Questions 19 and 24 from the current frame as context, again with Question 24’s answer from ground truth.
    \item Question 15 is answered, using the answer to Question 7 from the previous frame as context.
    \item Question 7 is answered, using the answer to Question 15 from the current frame as context.
    \item Question 8 is answered, with context including the answers to Questions 13, 47, 15, and 7 from the current frame.
    \item Question 43 is answered, using the answers to Questions 13 and 8 from the current frame as context.
    \item Finally, Question 50 is answered, with Question 43’s answer from the current frame included as context.
\end{enumerate}

\begin{table*}[tb!]
\centering
\small
\caption{\textbf{VQA performance of different VLMs with various integration strategies under Bench2Drive-VL}. \\ {Abbreviations: Imp. Obj. = Import Objects (Q19), T. Sign = Traffic Sign (Q15), S. Limit = Speed Limit (Q7), Col. Obj. = Collide Object (Q47), C. Lane = Change Lane (Q13), Brake = (Q8), A. Desc. = Action Description (Q43), A. Key = Action Keys (Q50).}\vspace{-2mm} \label{tab:model_comparison_VQA}
}
\resizebox{1.0\linewidth}{!}{
\begin{tabular}{l|cc|ccccccccc}
\toprule
\textbf{Model} & \textbf{Input} & \textbf{Chain} & \textbf{Imp. Obj.} & \textbf{T. Sign} & \textbf{S. Limit} & \textbf{Col. Obj.} & \textbf{C. Lane} & \textbf{Brake} & \textbf{A. Desc} & \textbf{A. Keys} \\ \midrule
\rowcolor{gray!10} DriveCommenter & & & 100.00 & 100.00 & 100.00 & 100.00 & 100.00 & 100.00 & 100.00 & 100.00 \\ \midrule
Qwen2.5VL-3B-Instruct & cam & yes & 53.62 & 75.47 & 27.58 & 5.21 & 81.81 & 62.94 & \textbf{69.69} & 46.51 \\
Qwen2.5VL-3B-Instruct & BEV & yes & 52.07 & 74.08 & 25.22 & 6.18 & 80.87 & 57.31 & 63.69 & 45.37 \\
Qwen2.5VL-3B-Instruct & cam & no & 50.37 & 78.76 & 24.64 & 4.82 & 75.04 & 57.67 & 58.07 & \textbf{51.03} \\
Qwen2.5VL-3B-Instruct & BEV & no & 53.45 & 72.06 & 30.33 & \textbf{8.31} & 66.27 & 50.39 & 52.63 & 44.42 \\ \midrule
Gemma3-4b-it & cam & yes & \textbf{60.52} & 21.24 & 19.34 & 7.29 & 25.07 & 30.32 & 15.27 & 13.47 \\
Gemma3-4b-it & BEV & yes & 59.62 & 29.42 & 18.24 & 8.12 & 25.18 & 22.98 & 14.89 & 11.15 \\
Gemma3-4b-it & cam & no & 55.82 & 17.03 & \textbf{100.00} & 6.02 & 7.11 & 27.81 & 59.69 & 49.64 \\
Gemma3-4b-it & BEV & no & 48.14 & 16.48 & 76.72 & 3.34 & 14.84 & 22.34 & 62.23 & 35.97 \\
\midrule
InternVL3-2B & cam & yes & 52.96 & 71.16 & 69.08 & 2.89 & \textbf{84.52} & 59.08 & 66.56 & 29.22 \\
InternVL3-2B & BEV & yes & 52.58 & \textbf{80.47} & 79.16 & 3.53 & 75.11 & \textbf{63.06} & 65.52 & 24.57 \\
InternVL3-2B & cam & no & 55.54 & 72.41 & 46.84 & 3.03 & 83.97 & 58.36 & 61.35 & 35.59 \\
InternVL3-2B & BEV & no & 54.41 & 70.54 & 44.59 & 2.82 & 80.55 & 54.41 & 66.22 & 32.46 \\
\bottomrule
\end{tabular}}
\vspace{-4mm}
\end{table*}

\textbf{Separation of CARLA Core and VLM.} To increase flexibility and simplify deployment, Bench2Drive-VL decouples the CARLA simulation from  VLM inference infrastructure. The VLM runs in an independent environment, communicating with the CARLA core via a web interface. This design allows the two components to operate on separate machines, accommodates differences in hardware requirements, and avoids the complexities of installing both CARLA and VLM in the same environment. Image data can be transmitted either as local paths (if on the same machine) or as base64-encoded strings (if across machines). New VLM models can be integrated simply by implementing a corresponding class.

During inference, the core CARLA module iterates through frames, generates prompts based on the configured reasoning graph, and sends these prompts along with visual inputs to the VLM module. The VLM's responses are returned and optionally stored for context propagation in subsequent frames.

\vspace{-2.5mm}\subsection{Action Module for Text to Control Signal} 

The action module of Bench2Drive-VL is responsible for translating the natural language answer from the VLM into control signals to execute in CARLA.

Among all questions, question 50 requires the VLM to select a direction key from [\texttt{FOLLOW\_LANE}, \texttt{CHANGE\_LANE\_LEFT}, \texttt{CHANGE\_LANE\_RIGHT}, \texttt{GO\_STRAIGHT}, \texttt{TURN\_LEFT}, \texttt{TURN\_RIGHT}, \texttt{DEVIATE\_LEFT}, \texttt{DEVIATE\_RIGHT}], and a speed key from [\texttt{KEEP}, \texttt{ACCELERATE}, \texttt{DECELERATE}, \texttt{STOP}]. Therefore, the final answer contains two key values, which are used by the action module to plan actions.

In some cases, the VLM does not output exactly two key values as expected. For example, some reasoning-focused models often produce a long reasoning procedure before giving the final answer, during which other key values might be mentioned. To extract the VLM's true intent in such cases, the reasoning module instructs the VLM in the prompt to place the final answer at the end, and the action module then takes the last occurring direction and speed key values as the final inference result of VLMs.

In other cases, due to limited capability, the VLM may fail to understand the question, resulting in no valid key values in the final answer. In such situations, if the direction key is missing, the default behavior is to continue along the current route; if the speed key is missing, the default is \texttt{KEEP}, i.e., to maintain the current speed.

Each time the VLM is invoked, the action module obtains new direction and speed key values. Based on these values, the action module modifies the waypoint list and target speed, which are then translated into concrete control signals by the lateral and longitudinal controllers. In frames where the VLM is not involved, the action module does not alter the current waypoint list, but inherits the most recent speed key value to maintain effective control of the ego vehicle.

\begin{table*}[tb!]
\centering
\small
\caption{\textbf{Planning performance of different VLMs with various integration strategies under Bench2Drive-VL}.\vspace{-2mm}\label{tab:model_comparison_behavior}}
\resizebox{1.0\linewidth}{!}{ 
\begin{tabular}{l|cc|cccc}
\toprule
\textbf{Model} & \textbf{Input} & \textbf{Chain} & \textbf{Driving Score $\uparrow$} & \textbf{Success Rate (\%) $\uparrow$} & \textbf{Efficiency $\uparrow$} & \textbf{Comfortness $\uparrow$} \\ \midrule
\rowcolor{gray!10} DriveCommenter & & & 100.00 & 100.00 & 154.91 & 12.48 \\ \midrule
Qwen2.5VL-3B-Instruct & cam & yes & 42.24 & 0.00 & 65.61 & 36.20 \\
Qwen2.5VL-3B-Instruct & BEV & yes & 48.05 & 0.00 & 84.79 & 38.63 \\
Qwen2.5VL-3B-Instruct & cam & no & \textbf{58.93} & \textbf{20.00} & 74.72 & 37.67 \\
Qwen2.5VL-3B-Instruct & BEV & no & 41.44 & \textbf{20.00} & 82.41 & 47.01 \\ \midrule
Gemma3-4b-it & cam & yes & 15.04 & 0.00 & 60.10 & 34.79 \\
Gemma3-4b-it & BEV & yes & 9.46 & 0.00 & 90.06 & 29.00 \\
Gemma3-4b-it & cam & no & 37.53 & 0.00 & \textbf{120.45} & 32.03 \\
Gemma3-4b-it & BEV & no & 49.52 & \textbf{20.00} & 108.13 & 23.05 \\ \midrule
InternVL3-2B & cam & yes & 33.00 & \textbf{20.00} & 48.36 & 65.48 \\
InternVL3-2B & BEV & yes & 29.73 & 0.00 & 79.40 & 51.33 \\
InternVL3-2B & cam & no & 29.92 & 0.00 & 51.18 & \textbf{77.45} \\
InternVL3-2B & BEV & no & 22.27 & 0.00 & 35.73 & 73.91 \\
\bottomrule
\end{tabular}}
\vspace{-4mm}
\end{table*}

\section{Experiments}\label{sec:experiments}

\subsection{Experiments Settings}

\noindent\textbf{Baselines.} We adapted three three commonly used  VLMs into Bench2Drive-VL as baselines: Qwen2.5VL-3B-Instruct\citep{qwen25report}, Gemma3-4b-it\citep{gemma3report}, and InternVL3-2B. 




\noindent\textbf{Input Configurations.} 
We evaluate both the camera input and BEV input setting supported by Bench2Drive-VL. We tested each model under both chain-of-thought (CoT) and non-CoT prompting strategies, with the VQA graph structure following~\autoref{fig:graph}. In the non-CoT mode, most questions were evaluated without context inheritance, except for Question 50, which inherits context from Question 43 because the action key values should be consistent with the VLM’s previous natural language description of the vehicle behavior.


\begin{figure*}[htbp]
    \centering
    \includegraphics[width=0.9\textwidth]{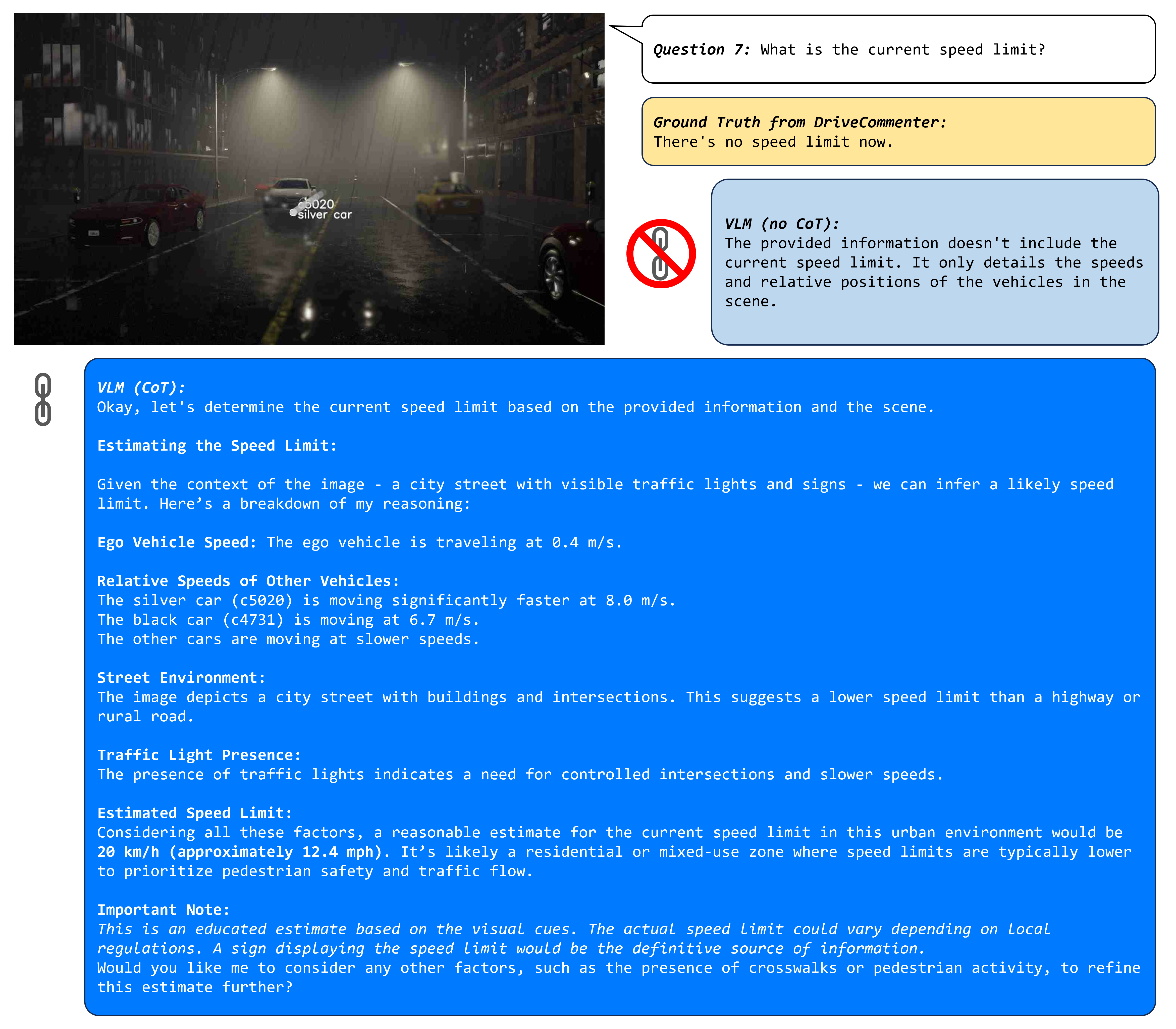}
    \caption{\textbf{Failure Case of CoT: Speed Limit Hallucation.}}
    \label{fig:speed_limit}
\end{figure*}

\vspace{-2.5mm}\subsection{Closed-Loop VQA and Planning Metrics}

Bench2Drive-VL's closed-loop evaluation consists of two parts: VQA and planning. The correctness of the VQAs is evaluated by a VQA evaluation module (utilizing DriveCommenter). The planning is evaluated by the standard Bench2Drive protocols.

\noindent\textbf{VQA Metrics}. Bench2Drive-VL leverages an LLM-as-Judge for VQA. As been widely acknowledged in the LLM community, direct LLM-based scoring is overly subjective~\citep{gu2025surveyllmasajudge, zheng2023judgingllmasajudgemtbenchchatbot}. Thus, Bench2Drive-VL adopts strict task-specific scoring rules. The evaluation LLM is used only for well-defined subtasks such as extracting objects in an answer, validating ordering, or interpreting control actions. For example: 

\begin{enumerate}[leftmargin=10pt, topsep=0pt, itemsep=1pt, partopsep=1pt, parsep=1pt]

\item Important Objects and Their Ordering.

No.19 question requires VLM to identify important objects in the scene and provide a ranked list. DriveCommenter first produces the ground-truth list of important objects and assigns each object two binary attributes: \texttt{is\_role} and \texttt{is\_dangerous}. Objects are ordered such that role objects have the highest priority, dangerous objects follow, and all other objects are ordered by distance.

Assume the ground-truth list contains $n$ objects $\mathbf{obj}_i$, indexed by importance ($i$ smaller means higher priority). Define a importance constant $R$ (usually $5$), The position weight is defined as
\begin{equation}
\mathbf{p}_i = \frac{N \cdot R}{R - 1} - i.
\end{equation}

The base weight is
\begin{equation}
\mathbf{b}_i =
\begin{cases}
3, & \text{if } \texttt{is\_role}(i) \text{ or } \texttt{is\_dangerous}(i),\\
1, & \text{otherwise}.
\end{cases}
\label{eq:basic_weights}
\end{equation}

The final importance weight is
\begin{equation}
\mathbf{w}_i = \mathbf{b}_i \cdot \mathbf{p}_i.
\end{equation}

Objects not appearing in ground truth are assigned a small penalty weight equal to the minimum position weight multiplied by $1/\texttt{EXTRA\_RATIO}$, where $\texttt{EXTRA\_RATIO}$ is set as $2$.

Ranking quality is measured by NDCG. Let the predicted list contain $k$ objects that overlap with the ground truth. The discounted cumulative gain is
\begin{equation}
\mathbf{DCG} = \sum_{i=1}^{k} \frac{\mathbf{w}_{r_i}}{\log_2(i+1)},
\end{equation}
and the ideal DCG is
\begin{equation}
\mathbf{IDCG} = \sum_{i=1}^{k} \frac{\mathbf{w}_{i}^{\text{ideal}}}{\log_2(i+1)}.
\end{equation}
Thus
\begin{equation}
\mathbf{NDCG} = \frac{\mathbf{DCG}}{\mathbf{IDCG}}.
\end{equation}

In addition, we compute a weighted F1-score:
\begin{subequations}
\label{eq:prf_block}
\begin{align}
\mathbf{P} &= \frac{\sum_{\text{TP}} \mathbf{w}_i}{\sum_{\text{TP}} \mathbf{w}_i + \sum_{\text{FP}} \mathbf{w}_i}, \\
\mathbf{R} &= \frac{\sum_{\text{TP}} \mathbf{w}_i}{\sum_{\text{TP}} \mathbf{w}_i + \sum_{\text{FN}} \mathbf{w}_i}, \\
\mathbf{F1} &= \frac{2 \cdot \mathbf{P} \cdot \mathbf{R}}{\mathbf{P} + \mathbf{R}}.
\end{align}
\end{subequations}

The final score for Question 19 is
\begin{equation}
\mathbf{score}_{19} = \mathbf{F1} \cdot \mathbf{NDCG}.
\end{equation}

\item Multi-object Reasoning Questions.

These questions require the VLM to identify a set of objects satisfying certain conditions and then provide object-specific descriptions. The evaluation consists of two components:

1. A weighted F1-score computed similarly to Equ.~\ref{eq:prf_block}, where extra-object penalties depend on the question type (e.g., small weight $0.25$ for harmless extra objects, or larger weight $1.5$ when extra objects indicate reasoning errors, e.g., collision prediction questions).

2. Object-wise correctness. For each object that appears in both the ground-truth and predicted lists, the evaluation model grades the object's answer, producing $\mathbf{s}_i$. The aggregated score is
\begin{equation}
\mathbf{S} = \frac{\sum_i \mathbf{s}_i \cdot \mathbf{b}_i}{\sum_i \mathbf{b}_i}.
\end{equation}

The final score is
\begin{equation}
\mathbf{score}_{\text{listed}} = \mathbf{F1} \cdot \mathbf{S}.
\end{equation}

\item Action Description and Control Keys.

In these questions, the VLM is required to select the correct action for the given situation. The final choice should include a direction key, represents the longitudinal control; as well as a speed key, which influences the lateral control. In our evaluation module, the critic model first parses the VLM's natural language answer to extract the implied control command. The extracted command is compared with the ground truth using an F1-score, then multiplied by a speed penalty:
\begin{equation}
\mathbf{score}_{\text{action}} = \mathbf{F1} \cdot \text{speed\_penalty}.
\end{equation}
The penalty is designed to reflect semantic severity (e.g., predicting \texttt{ACCELERATE} when the correct action is \texttt{STOP} yields the strongest penalty, while predicting \texttt{ACCELERATE} instead of \texttt{KEEP} is mildly favorable for efficiency).

\item Other Question Types.

For remaining question types, Bench2Drive-VL provides guideline prompts for the evaluation LLM. For example, answers exceeding the speed limit are penalized more heavily than those underestimating it; braking questions allow multiple acceptable answers when the ego vehicle is already stationary-such as \textit{"keep the current speed"} or \textit{"stop immediately"}.

\end{enumerate}

\noindent\textbf{Planning Metrics} follows Bench2Drive protocol~\cite{Bench2Drive}:

\begin{itemize}[leftmargin=10pt, topsep=0pt, itemsep=1pt, partopsep=1pt, parsep=1pt]
    \item \textbf{Driving Score}: average composed score based on route completion and violation penalties (e.g., collision penalty $0.6$, red-light or stop-sign violation $0.8$, timeout $0.7$).
    \item \textbf{Success Rate}: percentage of completion without penalty.
\end{itemize}

Bench2Drive also reports capability metrics (Merging, Overtaking, Emergency Brake, Give Way, Traffic Sign), driving efficiency (based on ego–traffic relative speed), and comfort metrics (percentage of trajectory satisfying acceleration and jerk thresholds).

\vspace{-2.5mm}\subsection{Quantitative Analysis of VLMs}
~\autoref{tab:model_comparison_behavior} and~\autoref{tab:model_comparison_VQA} demonstrate the results.

\begin{figure*}[htbp]
    \centering
    \includegraphics[width=0.9\textwidth]{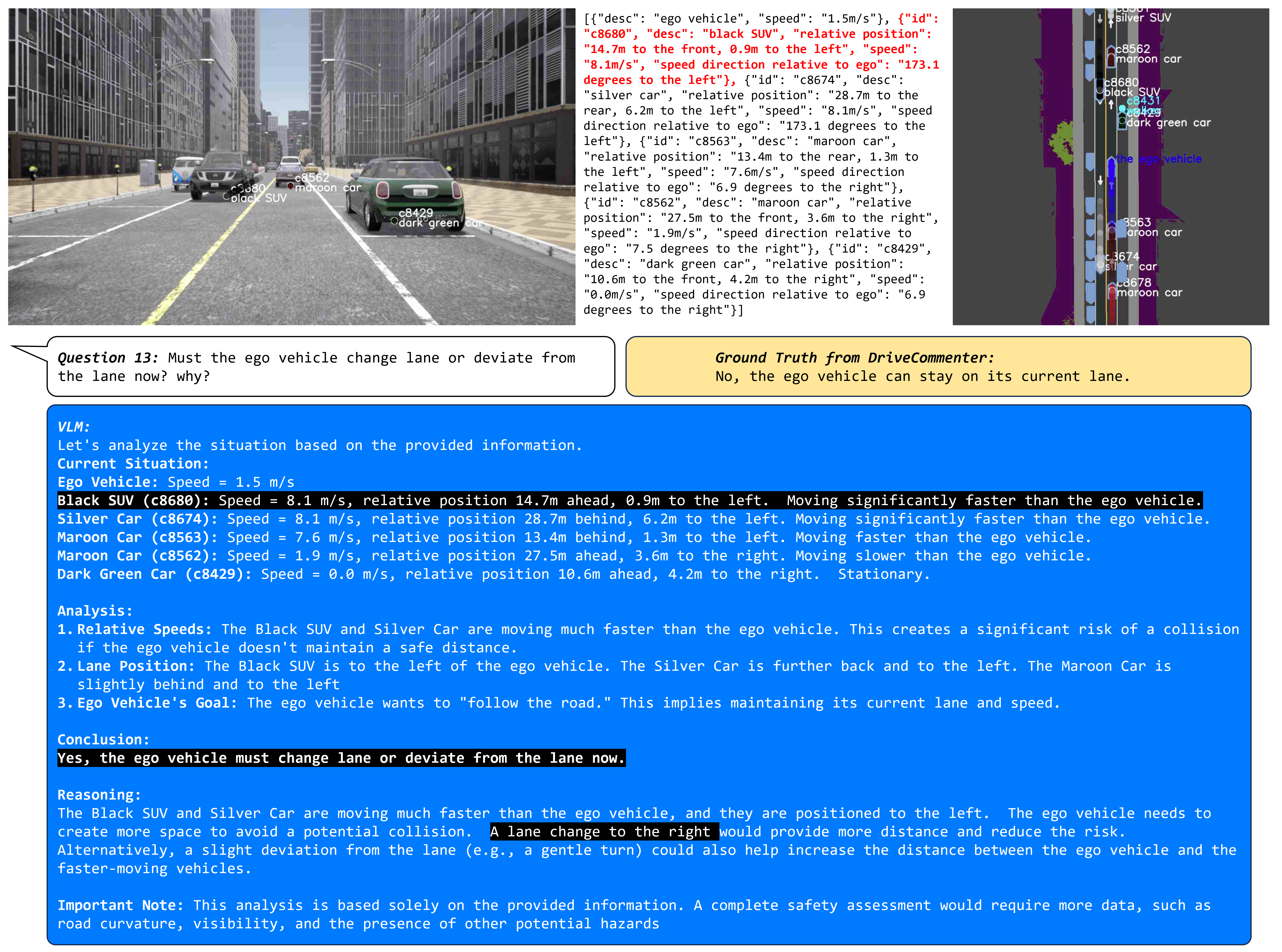}
    \vspace{-4mm}
    \caption{\textbf{Failure Case of BEV Map Input: Incorrect Lane Change.}}
    \vspace{-6mm}
    \label{fig:change_lane}
\end{figure*}

\noindent\textbf{Input Format Comparison.} The performance on camera and BEV inputs differs among VLMs. Camera inputs are more similar to the dataset most VLMs were pretrained on, thereby with better zero-shot understanding grounded in common sense. However, this familiarity can also induce hallucinations. In contrast, BEV inputs, while less appeared in VLM training sets and thus harder to interpret for some VLMs, provide more structured and explicit environmental representations.

\noindent\textbf{Inference Strategy Comparison.} Surprisingly, chain-of-thought (CoT) reasoning led to worse planning performance compared to non-CoT. This may be due to the models’ limited ability to handle long context, where accumulated context in CoT reasoning degrades inference quality and increases latency. Similar trends appear in VQA scores—Gemma3-4b-it, for instance, suffers from hallucinations in speed limit questions under CoT, often inferring non-existent limits based on irrelevant context.

\noindent\textbf{Model Comparison.} Among the three models tested, Qwen2.5VL-3B-Instruct outperformed InternVL3-2B and Gemma3-4b-it in planning. Gemma3-4b-it's performance was hindered by its tendency toward long, complex reasoning, often resulting in hallucinated lane changes and unsafe maneuvers. In contrast, Qwen2.5VL-3B-Instruct and InternVL3-2B adopted more conservative strategies, who rarely try special moves. InternVL3-2B was particularly cautious, often driving extremely slowly throughout a scenario, which, while safe, significantly sacrificed efficiency and would be impractical in real-world applications.

\vspace{-2.5mm}\subsection{Faiure Case Studies}

\noindent\textbf{Speed Limit Hallucination.}  
Under CoT, VLMs often predicts non-existent speed limits based on irrelevant context (\autoref{fig:speed_limit}). In non-CoT mode, predictions better match scene.

\noindent\textbf{Incorrect Lane Change.}  With BEV inputs (\autoref{fig:change_lane}), a black vehicle approached rapidly in the left lane. Gemma3-4b-it attempted a right lane change despite the stationary vehicle in the right lane, resulting in potential collision. Qwen2.5VL-3B-Instruct and InternVL3-2B consistently adopt conservative strategies, avoiding unsafe maneuvers.

\begin{figure*}[!t]
    \centering
    \includegraphics[width=0.9\textwidth]{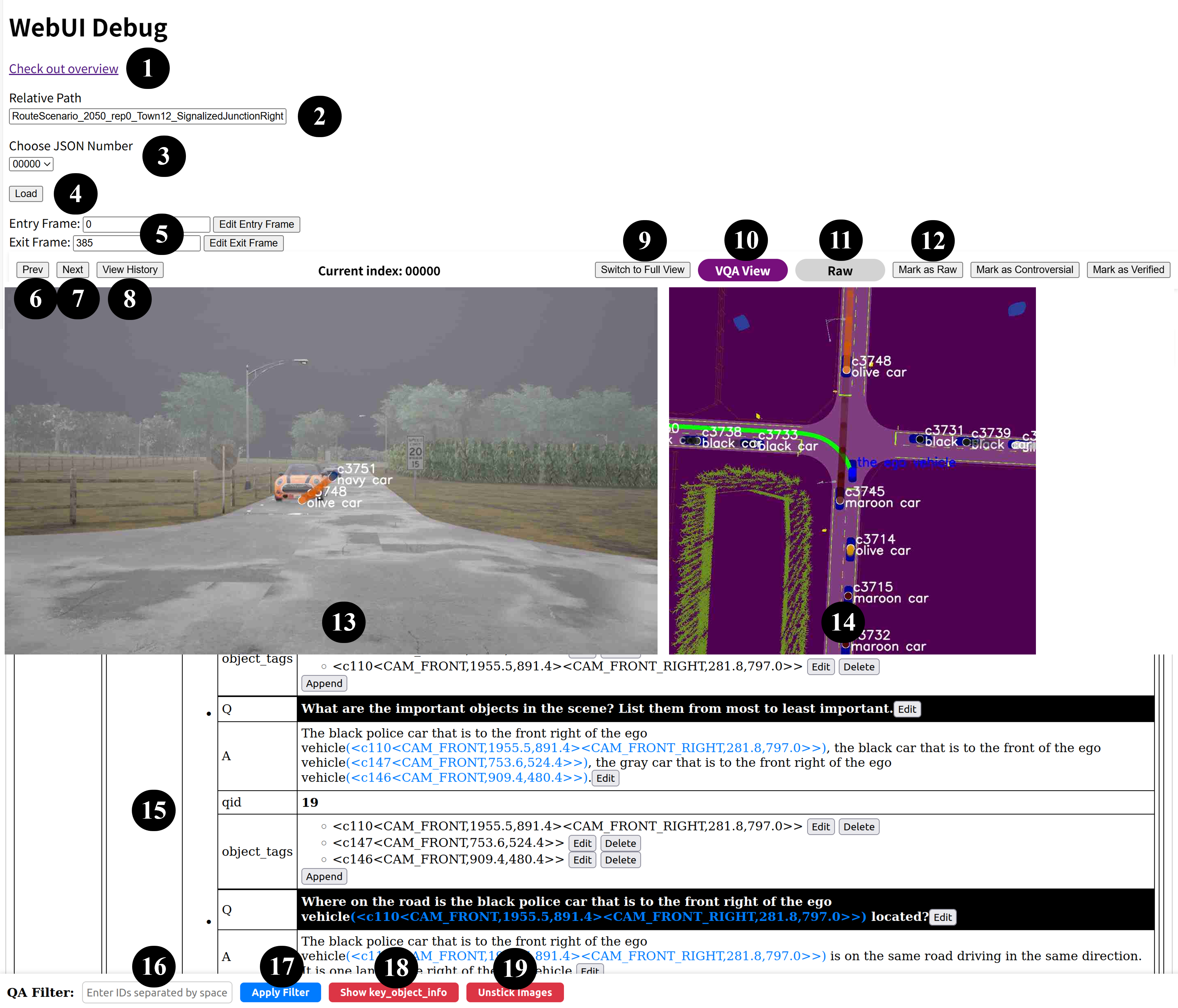}
    \caption{\textbf{Main Interface of DLVis.}}
    \label{fig:screenshot1}
\end{figure*}

\section{Tools for Developing VLM4AD Datasets and Models}
The development toolkit open sourced by Bench2Drive-VL is DriveLangVis (DLVis), a web-based tool for visualizing and editing data for developing VLM4AD datasets and CoT models, including both original scene annotations and VQA outputs generated by DriveCommenter. Since inference and evaluation results of Bench2Drive-VL shares the same format with Bench2Drive-VL dataset, DLVis can be used to browse them as well. After configuring input paths via a YAML file, users can launch DLVis locally and interactively browse driving scenarios through two main views: a compact VQA-only mode and a comprehensive full mode showing images, BEV (bird’s-eye view) annotations, and language outputs. DLVis supports frame-by-frame navigation, in-browser editing of questions and answers, marking of controversial or verified entries, and history tracking of all edits. It also provides advanced functionalities such as filtering by question ID or keyword, setting frame ranges to exclude noisy segments, and pinning reference images for comparison across frames. 

\begin{figure*}[!t]
    \centering
    \includegraphics[width=0.9\textwidth]{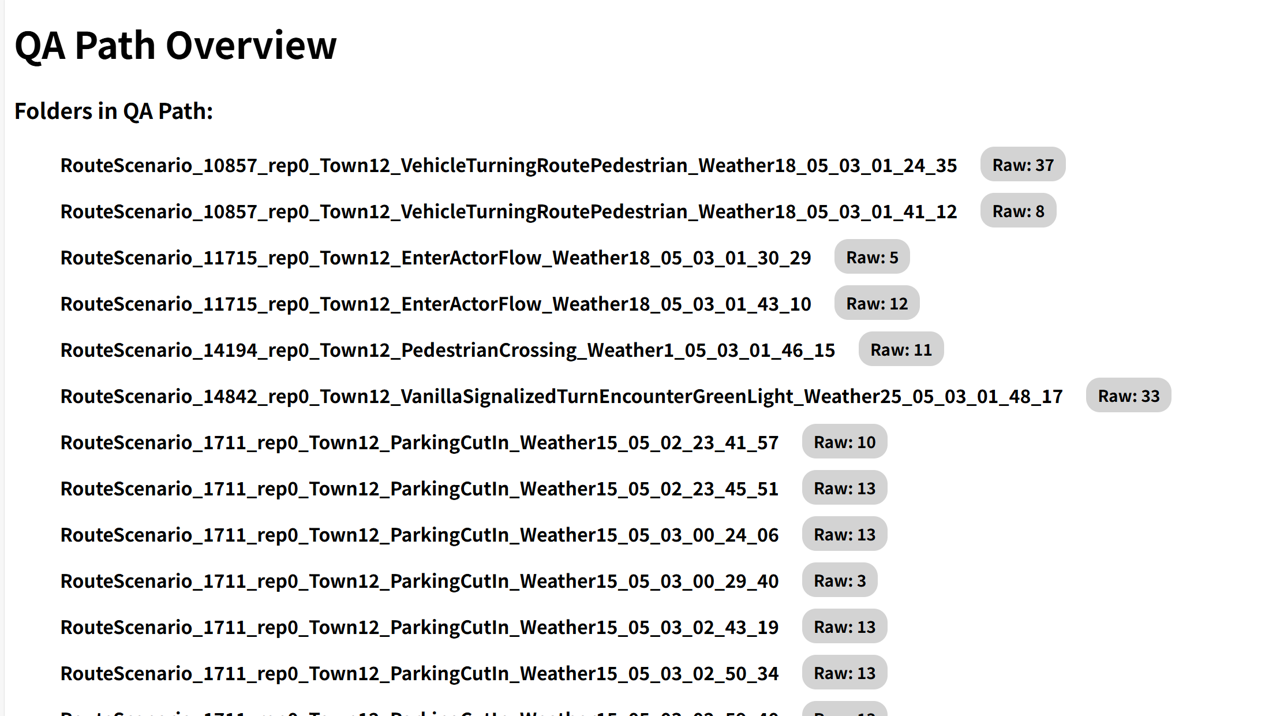}
    \caption{\textbf{Data Overview Interface of DLVis.}}
    \label{fig:screenshot2}
\end{figure*}

\begin{figure*}[!t]
    \centering
    \includegraphics[width=0.9\textwidth]{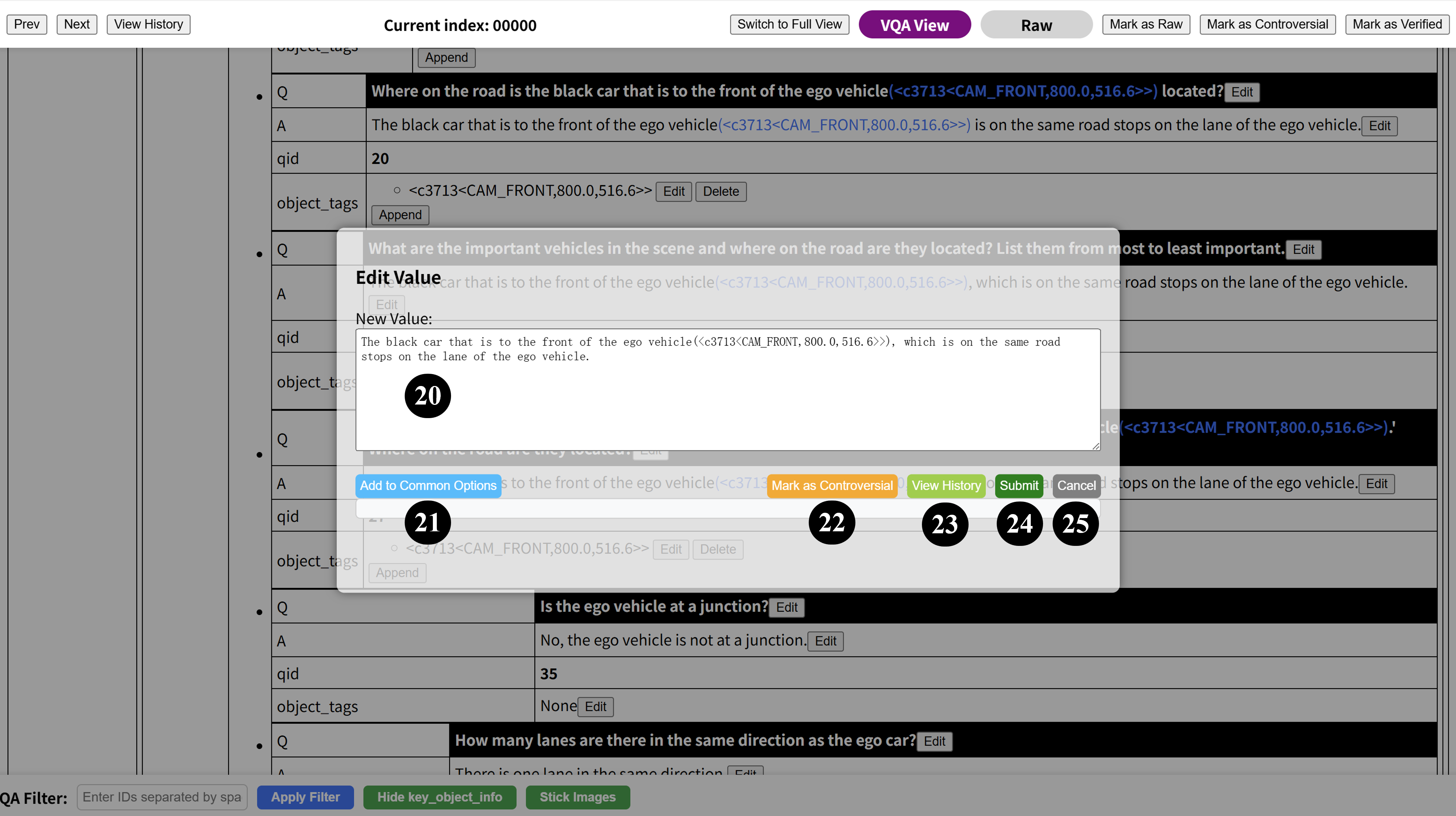}
    \caption{\textbf{Editing Interface of DLVis.}}
    \label{fig:screenshot3}
\end{figure*}

\autoref{fig:screenshot1} shows the main interface of our visualization tool. The functions of labeled components are:

\begin{enumerate}
    \item A button to jump to the data overview interface shown in \autoref{fig:screenshot2}. This page lists the annotation status of the dataset, including the labeling status of each frame in every scenario subset ("raw" (unprocessed), "controversial", or "verified"). This helps the user keep track of the overall progress.
    \item A text field in which the scenario to show below is chosen.
    \item A text field to select the frame to show below.
    \item A button to load the frame and scenario selected above.
    \item The text field for editing the entry and the exit frame of selected scenario. Sometimes, the data at the beginning or end of a scenario is problematic when the data is annotated from the static Bench2Drive dataset. If the user need to discard them, make the adjustments here. Only the frame interval \text{[entry frame, exit frame)} will be preserved for further processing.
    \item A button for navigating to the previous frame.
    \item A button for navigating to the next frame.
    \item A button for showing all modification records of this frame's data.
    \item A button for changing views. The VQA View only shows VQAs and the information of key objects of the current frame, while the Full View shows all information, including bounding boxes and other detailed metadata of all objects in the current frame.
    \item An indicator that shows the current view the tool is in.
    \item An indicator that shows the data status of the current frame ("raw", "controversial", or "verified").
    \item Buttons that mark the current frame's data status.
    \item The place showing image type 1.
    \item The place showing image type 2. The images shown in these two places can be freely configured. In this screenshot, the annotated RGB front camera image is shown on the left, while the annotated BEV is shown on the right.
    \item The place showing editable data in the current frame.
    \item The test field that is used to filter VQA. The user can enter the \texttt{qid}s of the VQAs that are expected to be shown. If it is left blank, all VQAs are shown.
    \item The button that applies the filter configuration on the left.
    \item The button that switches the showing status of the key objects' information.
    \item The button that switches the rendering mode of images. They can be whether stuck on the top of the page or not.
    \item  Below are the editing interface in \autoref{fig:screenshot3}.  The text field that is used to edit the value of selected data. If the user clicks the edit button in (15), this window will pop up.
    \item The button to add the current content in the text field to the common option list of this question. After that, this option will always be shown when the user is editing a question with the same \texttt{qid} as this one.
    \item The button to mark this data as controversial. If any data in a certain frame is controversial, this frame is controversial.
    \item The button to view the modification history of this data.
    \item The button to exit the editing interface.
\end{enumerate}

The visualization tool's detailed documentation can be found in \url{https://thinklab-sjtu.github.io/Bench2Drive-VL/docs/tutorial/visualization}.

\section{Conclusion}

In this work, we presented \textbf{Bench2Drive-VL}, a comprehensive framework for closed-loop, question-driven evaluation of vision-language model (VLM) based autonomous driving agents. Building upon Bench2Drive~\citep{Bench2Drive}, our extension integrates \textbf{DriveCommenter}, an adaptive expert agent capable of generating behavior-consistent VQA annotations in real time across diverse and challenging driving scenarios. Bench2Drive-VL further introduces a unified communication protocol connecting CARLA simulations with modern VLMs, flexible graph-based chain-of-thought reasoning interfaces, and an integrated suite of developer tools for visualization, debugging, and deployment.

Through extensive evaluations, we demonstrated that Bench2Drive-VL enables holistic assessment of VLM-based agents, encompassing planning, perception, reasoning, and control. Our experiments reveal that medium-scale VLMs, while capable of understanding multimodal inputs, still exhibit hallucinations and over-conservative behaviors under long-context scenarios, highlighting the gap between current models and usable performance. 

Overall, Bench2Drive-VL offers a scalable, reproducible, and extensible platform for evaluating and advancing VLM4AD. We believe that Bench2Drive-VL will serve as a critical benchmark and development hub for the community, facilitating the design of more capable, safe, and interpretable VLM-based driving agents.

\bibliographystyle{IEEEtran}
\bibliography{egbib}

\end{document}